\documentclass[journal]{IEEEtran}

\usepackage{cite}

\ifCLASSINFOpdf
\usepackage[pdftex]{graphicx}
\else
\fi

\usepackage{amsmath}
\usepackage{multirow}
\usepackage[linesnumbered,lined,ruled,commentsnumbered]{algorithm2e}
\usepackage[switch]{lineno}

\begin{document}

\title{Effective Sequential Classifier Training for SVM-Based Multitemporal Remote Sensing Image Classification}

\author{Yiqing~Guo,~\IEEEmembership{Student Member,~IEEE,}
        Xiuping~Jia,~\IEEEmembership{Senior Member,~IEEE,}
        and~David~Paull,~\IEEEmembership{}

\thanks{Y. Guo and X. Jia are with the School of Engineering and Information Technology, The University of New South Wales, Canberra Campus, Canberra, ACT 2600, Australia (e-mail: Yiqing.Guo@student.adfa.edu.au; X.Jia@adfa.edu.au).  \textit{(Corresponding author: Xiuping Jia.)} }
\thanks{D. Paull is with the School of Physical, Environmental and Mathematical Sciences, The University of New South Wales, Canberra Campus, Canberra, ACT 2600, Australia (e-mail: D.Paull@adfa.edu.au)}

}

\maketitle

\begin{abstract}
The explosive availability of remote sensing images has challenged supervised classification algorithms such as Support Vector Machines (SVM), as training samples tend to be highly limited due to the expensive and laborious task of ground truthing. The temporal correlation and spectral similarity between multitemporal images have opened up an opportunity to alleviate this problem. In this study, a SVM-based Sequential Classifier Training (SCT-SVM) approach is proposed for multitemporal remote sensing image classification. The approach leverages the classifiers of previous images to reduce the required number of training samples for the classifier training of an incoming image. For each incoming image, a rough classifier is firstly predicted based on the temporal trend of a set of previous classifiers. The predicted classifier is then fine-tuned into a more accurate position with current training samples. This approach can be applied progressively to sequential image data, with only a small number of training samples being required from each image. Experiments were conducted with Sentinel-2A multitemporal data over an agricultural area in Australia. Results showed that the proposed SCT-SVM achieved better classification accuracies compared with two state-of-the-art model transfer algorithms. When training data are insufficient, the overall classification accuracy of the incoming image was improved from 76.18$\%$ to 94.02$\%$ with the proposed SCT-SVM, compared with those obtained without the assistance from previous images. These results demonstrate that the leverage of \emph{a priori} information from previous images can provide advantageous assistance for later images in multitemporal image classification.
\end{abstract}


\IEEEpeerreviewmaketitle

\section{Introduction}\label{sec: introduction}

\IEEEPARstart{L}{and} cover information extracted from remote sensing images has enabled timely monitoring of land cover dynamics at a large spatial scale \cite{verbesselt2010phenological,gomez2016optical}. For the generation of land cover maps, supervised classification algorithms have been widely adopted \cite{richards2013remote}. Ground reference data are required by these algorithms for the training of classifiers \cite{richards2013remote}.

However, traditional supervised classification algorithms such as the Support Vector Machines (SVM) have been challenged by the explosive availability of remote sensing images. While collecting a sufficient number of training samples is critical to achieve satisfactory classification results, it is often labour- and time-consuming and sometimes practically unfeasible \cite{bruzzone2014review,shao2012comparison,bahirat2012novel,demir2013classification,morgan2002learning}. The problem is more severe for agricultural areas where the accuracy of ground reference data can be quickly outdated, as a result of the frequent land cover changes and spectral characteristic shifts caused by crop phenology and anthropogenic practices such as irrigation and harvesting \cite{sakamoto2005crop}. As a result, training data sufficiency is often hard to be guaranteed for every image.

Temporal correlation and spectral similarity between multitemporal images have opened up an opportunity to alleviate the \emph{small sample size} problem. When an incoming image needs to be classified, existing knowledge provided by previous images can be utilized as \emph{a priori} information. However, it cannot be directly applied to the incoming image, because class data of different images may follow different probabilistic distributions in the feature space. This phenomenon is often referred to as cross-image dataset shifts \cite{moreno2012unifying}. The shifts can either originate from changes in the nature of land surface properties, for example, induced by the phenology of vegetation, or from the background noise caused by varied acquisition and atmospheric conditions \cite{vanonckelen2013effect,xu2016thin,xu2016cloud} and inconsistent sun-target-senor geometries \cite{zhao2013bidirectional,guo2017superpixel}. Therefore, appropriate domain-adaptation strategies need to be applied to tackle the dataset shifts \cite{tuia2016domain,tuia2011using,yang2016domain,bruzzone2009toward,banerjee2015novel,matasci2015semisupervised}.

Domain adaptation is an emerging technique that is able to accommodate the dataset shifts among images. The technique aims to adapt \emph{a priori} information extracted from previous images to an incoming image that has shifted spectral characteristics. Compared with a complete retraining of the incoming image, the required number of training samples can be considerably reduced if the assistance from previous images is leveraged \cite{bahirat2012novel}. Several domain adaptation algorithms have been proposed to accommodate the dataset shifts in multitemporal image classification. For example, the change-detection-driven algorithms \cite{demir2013classification,demir2013updating} train a classifier for an incoming image with the help of the unchanged samples from a previous image. A change detection step is firstly implemented to identify unchanged pixels, whose labels are then directly transferred to the incoming image. However, as erroneous labels may be transferred as well, the classification error will then propagate from the previous image to the incoming image and accumulate when the algorithm is applied progressively to multiple incoming images. The Bayesian-based algorithms \cite{bahirat2012novel} adapt a Bayesian classifier to an incoming image with the statistical parameters calculated from the previous image as an \emph{a priori} estimate. Then a retraining procedure is used to adjust the \emph{a priori} estimate, making it right for the incoming image. However, the \emph{a priori} estimate is made with only one previous image instead of a set of previous images, so the temporal trajectory of class spectral characteristics, which is useful information to make a more accurate \emph{a priori} estimation, is not extracted and utilized in the algorithm.

This study provides a SVM-based sequential classifier training (SCT-SVM) approach for multitemporal remote sensing image classification. With the assistance from a set of previous classifiers, this approach is able to effectively train classifiers for an incoming image at the same location. This approach can be applied progressively to sequential image data, with only a small number of training samples being required from each incoming image. Previous studies have demonstrated that parameters of previous classifiers can be updated for an incoming image in a semisupervised manner \cite{bahirat2012novel,ghoggali2008genetic,bovolo2008novel,yuan2015semi,chen2009semi}. In these studies, labelled pixels from previous images and unlabelled pixels from the incoming image are utilized in updating Bayesian \cite{bahirat2012novel}, Gaussian-Process (GP) \cite{chen2009semi}, or SVM classifiers \cite{bovolo2008novel} for multitemporal remote sensing image classification and/or change detection. Specifically, the class data of the incoming image are matched to those of previous images based on their spectral similarities. This method assumes that the labelled dataset of previous images is stored and accessible. Class separability needs to be adequate to avoid a risk of matching to a wrong class. Different from previous studies, raw data (pixels and their corresponding labels) of previous images are not required by the proposed approach. Instead, only the classifier-level data (classifier parameters of a set of previous classifiers) are needed. This also leads to better computational efficiency. In the present study, a set of Sentinel-2A multitemporal images over an agricultural area in Australia has been used to demonstrate the effectiveness of the proposed approach.

\section{Method}\label{sec: method}

\subsection{Proposed Framework}
When an incoming image needs to be classified, the \emph{small sample size} problem often arises if training samples are difficult to collect. In order to mitigate this problem, we propose to leverage \emph{a priori} knowledge gained from the classification of previous images at the same location. The assistance from previous images is provided at the classifier-level instead of the data-level. Specifically, the parameters of classifiers instead of raw data (i.e., pixel values and/or their corresponding labels) from the previous images are used. The proposed approach does not make assumptions on the way previous classifiers are trained and how many training samples are used, except that they are reliable in classifying their respective images.

According to the sequence of sensing date, previous images and their classifiers are indexed $t-1$, $t-2$, $\cdots$, $t-N$, where $N$ is the total number of previous images, and the incoming image and its classifier are indexed $t$. Mathematically there is no fixed requirement of $N$ in the proposed method. In practice, $N$ is often limited and the selection of $N$ is also case-dependent. Generally, a larger number of previous classifiers that are close to the sensing date of the incoming image is better. Other factors, such as how good each previous classifier is and how long the period of the previous images cover, need to be considered for the selection of $N$.

\begin{figure*}[!htb]
\centering
\includegraphics[width=6.3in]{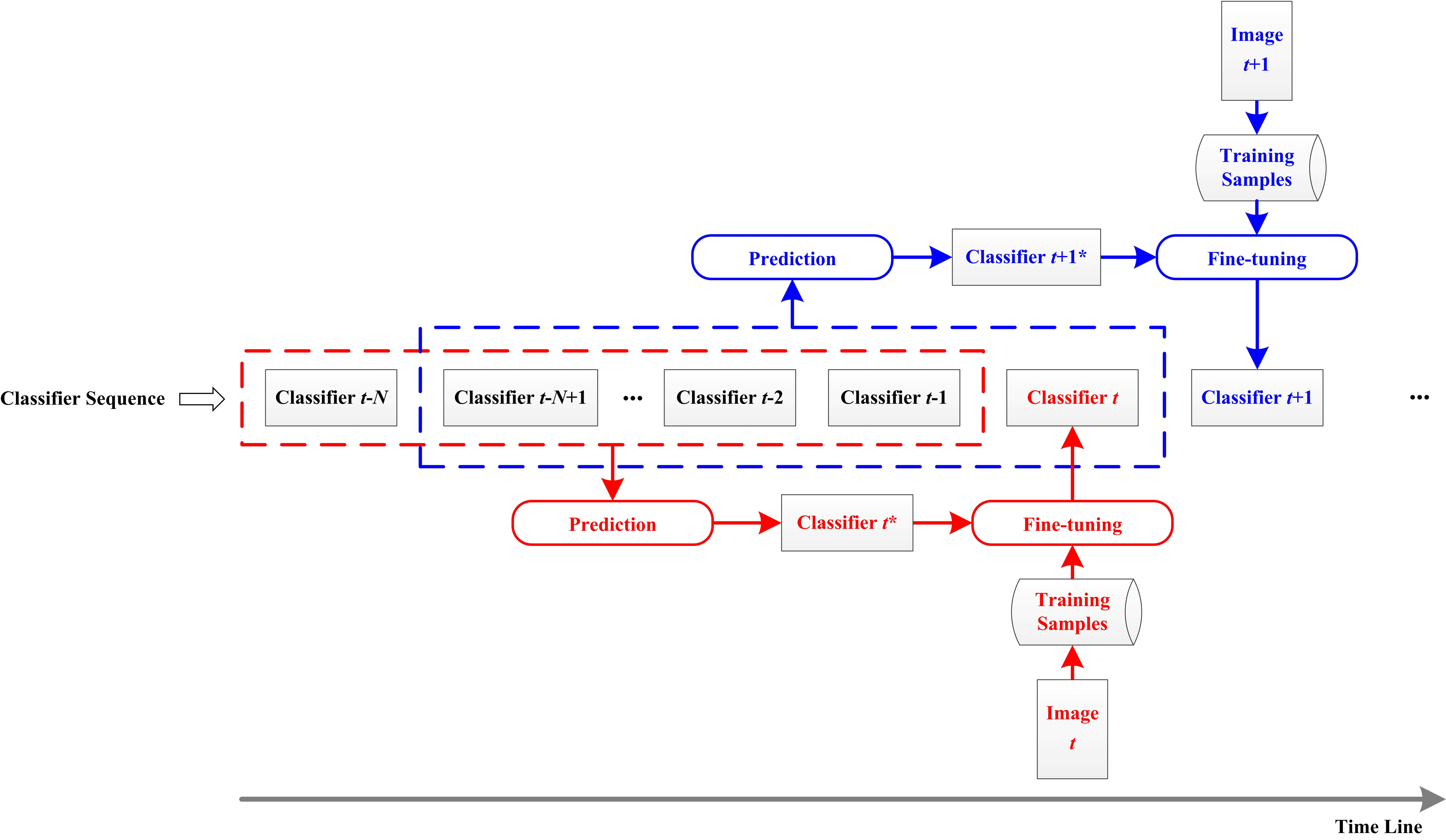}
\caption{The framework of the proposed sequential classifier training approach for multitemporal remote sensing image classification.}\label{Fig: multiple}
\end{figure*}

Consider that a classifier needs to be determined for Image $t$. We propose a sequential classifier training approach that consists of two steps: prediction and fine-tuning, as shown in Fig. \ref{Fig: multiple}. In the prediction step, a rough classifier (Classifier $t^*$) is firstly estimated based on a set of previous classifiers (Classifiers $t-1$, $t-2$, $\cdots$, $t-N$). In the fine-tuning step, the predicted classifier (Classifier $t^*$) is adjusted and finalized as Classifier $t$ with current training samples. Because the fine-tuning is based on the predicted classifier, compared with the direct training of a classifier for Image $t$ without any ancillary information, the required number of training samples can be smaller. After Classifier $t$ is determined, it can be included into the set of previous classifiers, while the oldest classifier (Classifier $t-N$) may be removed. The updated set is then used as \emph{a priori} knowledge to assist the training of the next classifier (Classifier $t+1$).

The proposed approach is named as SCT-SVM as it is based on the well-established classification algorithm of Support Vector Machines (SVM). In a $m$-dimensional feature space, a linear SVM classifier $f\left( \mathbf{x} \right)$ for a binary classification problem can be expressed as \cite{mountrakis2011support}:
\begin{equation}
f(\mathbf{x})={{\mathbf{w}}^{\rm{T}}}\mathbf{x}+b,
\end{equation}
where $\mathbf{x}={{\left( {{x}_{1}}\text{ }{{x}_{2}}\text{ }\cdots \text{ }{{x}_{m}} \right)}^{\text{T}}}$ is the feature vector; $\mathbf{w}={{\left( {{w}_{1}}\text{ }{{w}_{2}}\text{ }\cdots \text{ }{{w}_{m}} \right)}^{\text{T}}}$ is the weight vector and $b$ is the bias factor. For convenience, we stack the classifier parameters $\mathbf{w}$ and $b$ into a single vector $\mathbf{p}={{\left( {{w}_{1}}\text{ }{{w}_{2}}\text{ }\cdots \text{ }{{w}_{m}}\text{ }b \right)}^{\text{T}}}={{\left( {{\mathbf{w}}^{\text{T}}}\text{ }b \right)}^{\text{T}}}$. Because each classifier $f(\mathbf{x})$ can be uniquely determined by its parameters $\mathbf{p}$, the terms \emph{classifier} and \emph{classifier parameters} will be used synonymously in the following. The binary SVM classifier can handle the multi-class case by applying the \emph{one-against-one} or \emph{one-against-the-rest} strategies \cite{richards2013remote}.

\begin{figure}[!htb]
\centering
\includegraphics[width=3in]{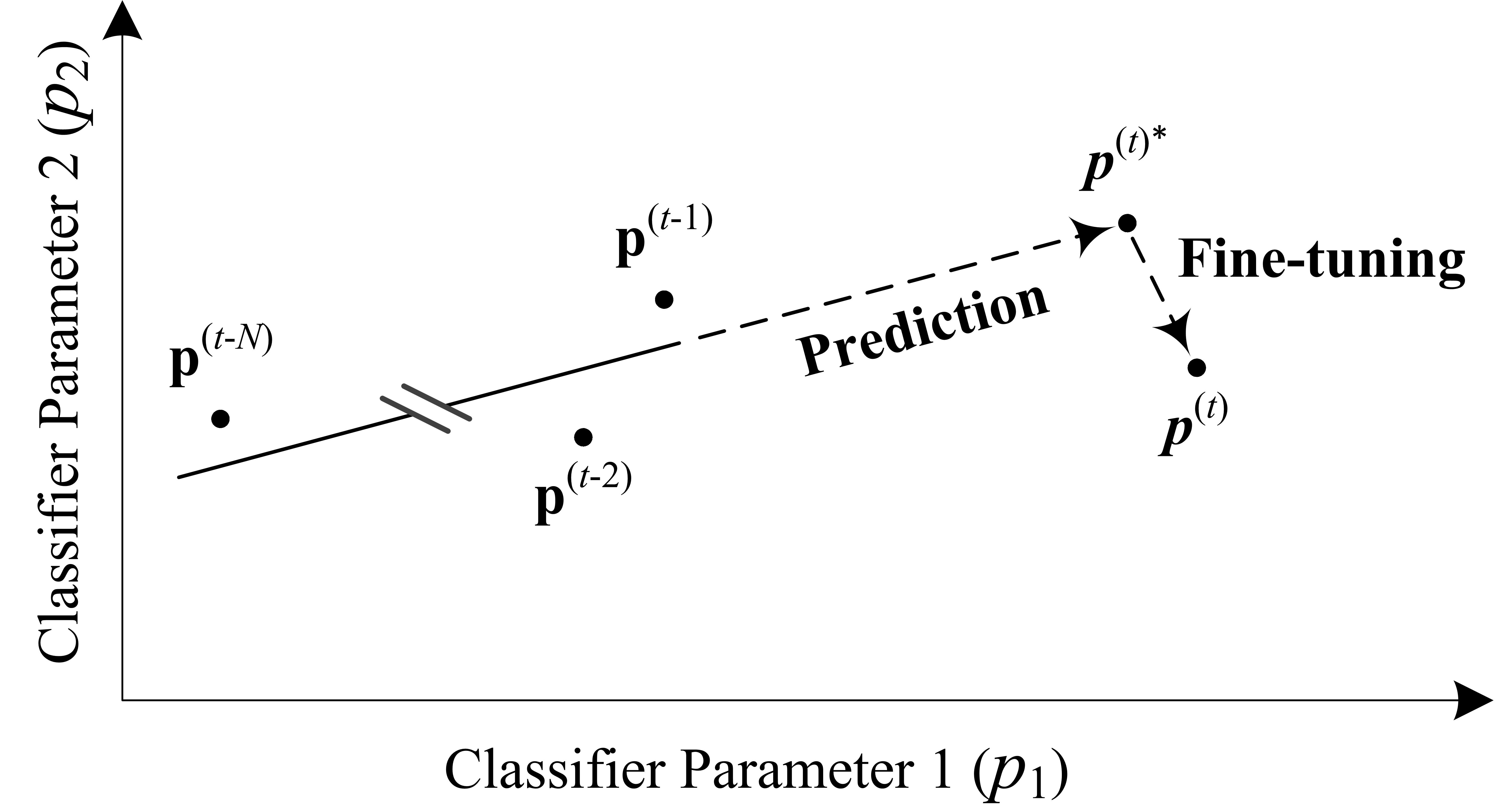}
\caption{Illustration of the two step classifier training in the classifier parameter space.}\label{Fig: framework}
\end{figure}

Figure \ref{Fig: framework} illustrates the classifier parameter space defined in two dimensions (e.g., $p_1$ and $p_2$), where the prediction/fine-tuning steps are detailed. Different classifiers often appear at different points in this parameter space, due to the temporal variations among multitemporal images. The temporal trend of previous classifiers ($\mathbf{p}^{(t-1)}$, $\mathbf{p}^{(t-2)}$, $\cdots$, $\mathbf{p}^{(t-N)}$) is firstly utilized to predict a rough classifier ($\mathbf{p}^{(t)*}$) for the current image (Image $t$), and it is then fine-tuned into an updated position ($\mathbf{p}^{(t)}$). The prediction and fine-tuning steps will be detailed below in the following Sections. \ref{ssec: prediction} and \ref{ssec: fine-tuning}, respectively.

\subsection{Classifier Prediction}\label{ssec: prediction}
\begin{figure}[!htb]
\centering
\includegraphics[width=2.8in]{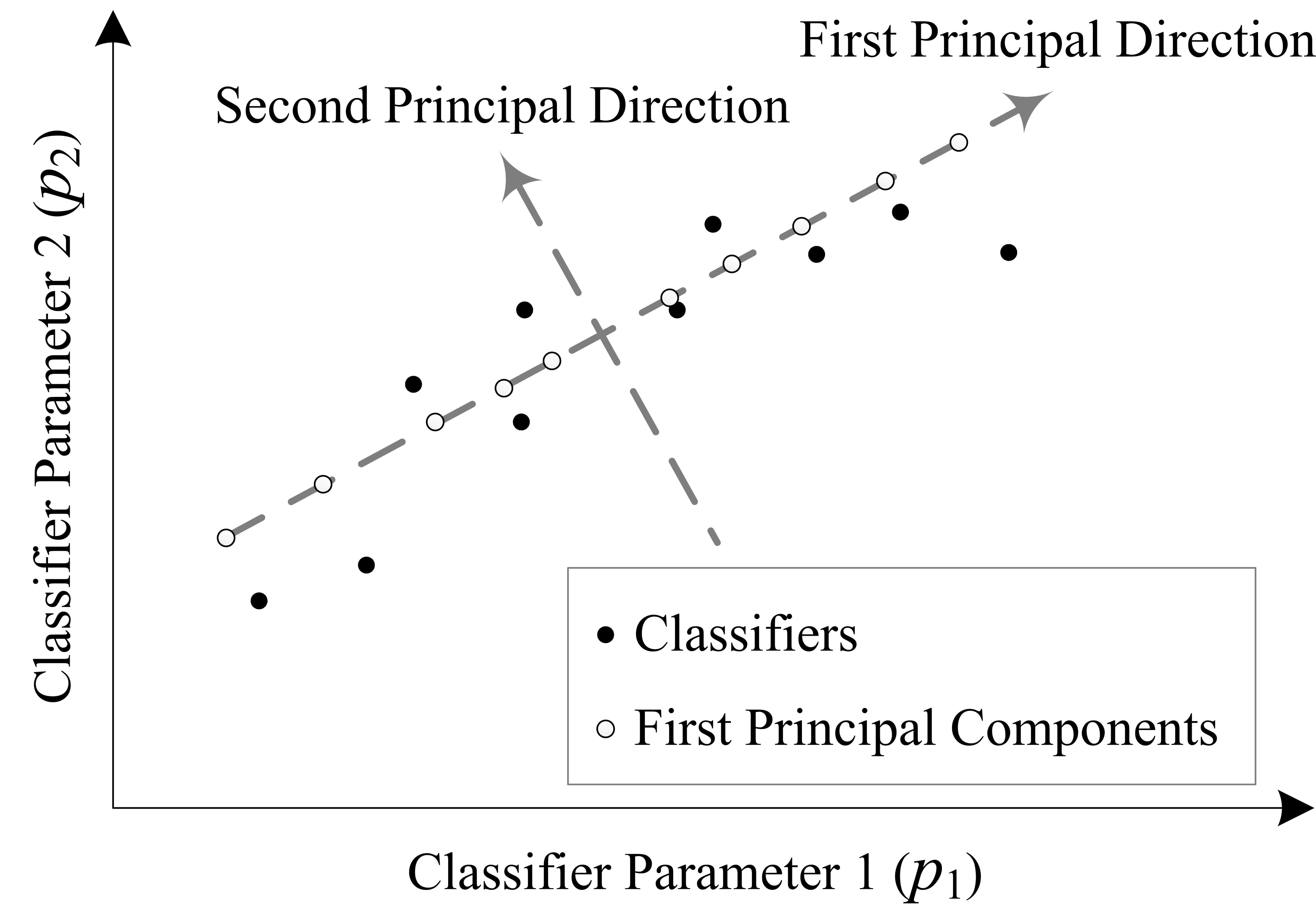}
\caption{Principal component analysis in the classifier parameter space.}\label{Fig: first}
\end{figure}

The aim of classifier prediction is to determine a rough classifier for the current image (Image $t$). The prediction is based on the temporal trend of a set of previous classifiers. The trend can be due to the background variations under different acquisition conditions or originated from changes in the nature of land surface properties such as the phenology of crops and forests, or both. The former factor causes random and unpredictable changes. The latter factor induces relatively gradual changes, due to the smooth transitions of class data over time. We propose to perform a principal component transform, and use the first transformed component to extract the overall temporal trend and suppress the background effects. As shown in Fig. \ref{Fig: first}, the first principal component direction captures the maximum temporal dynamics of previous classifiers.

The classifier prediction is conducted with the following procedure. Firstly, the previous classifiers are mean-centralized in the classifier parameter space (Fig. \ref{Fig: first}):
\begin{equation}
{{\mathbf{\bar{p}}}^{\left( i \right)}}={{\mathbf{p}}^{\left( i \right)}}-\frac{1}{N}\sum\limits_{j=t-1}^{t-N}{{{\mathbf{p}}^{\left( j \right)}}}\text{,}\quad i=t-1\text{, }t-2\text{, }\cdots \text{, }t-N\text{,}
\end{equation}
where ${{\mathbf{p}}^{\left( i \right)}}$ and ${{\mathbf{\bar{p}}}^{\left( i \right)}}$ are the classifiers before and after mean-centralization, respectively.

Then a principal component transform is applied to transform the mean-centered classifiers from the classifier parameter space into the principal component space:
\begin{equation}
{{\mathbf{\bar{q}}}^{\left( i \right)}}=\mathbf{G}{{\mathbf{\bar{p}}}^{\left( i \right)}}\text{,}\quad i=t-1\text{, }t-2\text{, }\cdots \text{, }t-N,
\end{equation}
where ${{\mathbf{\bar{q}}}^{\left( i \right)}}={{\left( {{{\bar{q}}}_{1}}^{\left( i \right)}\text{, }{{{\bar{q}}}_{2}}^{\left( i \right)},\text{ }\cdots \text{, }{{{\bar{q}}}_{m+1}}^{\left( i \right)} \right)}^{\text{T}}}$ are the transformed classifiers with ${{\bar{q}}_{1}}^{\left( i \right)}$ being their first principal components, and $\mathbf{G}$ is a $(m+1)$-by-$(m+1)$ matrix that provides the transformation coefficients.

The transformation matrix $\mathbf{G}$ is calculated by decomposing the matrix $\mathbf{\bar{P}}=\left( {{{\mathbf{\bar{p}}}}^{\left( t-1 \right)}}\text{ }{{{\mathbf{\bar{p}}}}^{\left( t-2 \right)}}\text{ }\cdots \text{ }{{{\mathbf{\bar{p}}}}^{\left( t-N \right)}} \right)^{\text{T}}$ with singular value decomposition:
\begin{equation}
\mathbf{\bar{P}}=\mathbf{U}\Lambda {{\mathbf{G}}^{\text{T}}},
\end{equation}
where $\mathbf{U}$ and $\Lambda$ are $N$-by-$N$ and $N$-by-$(m+1)$ matrices, respectively.

\begin{figure}[!htb]
\centering
\includegraphics[width=3.4in]{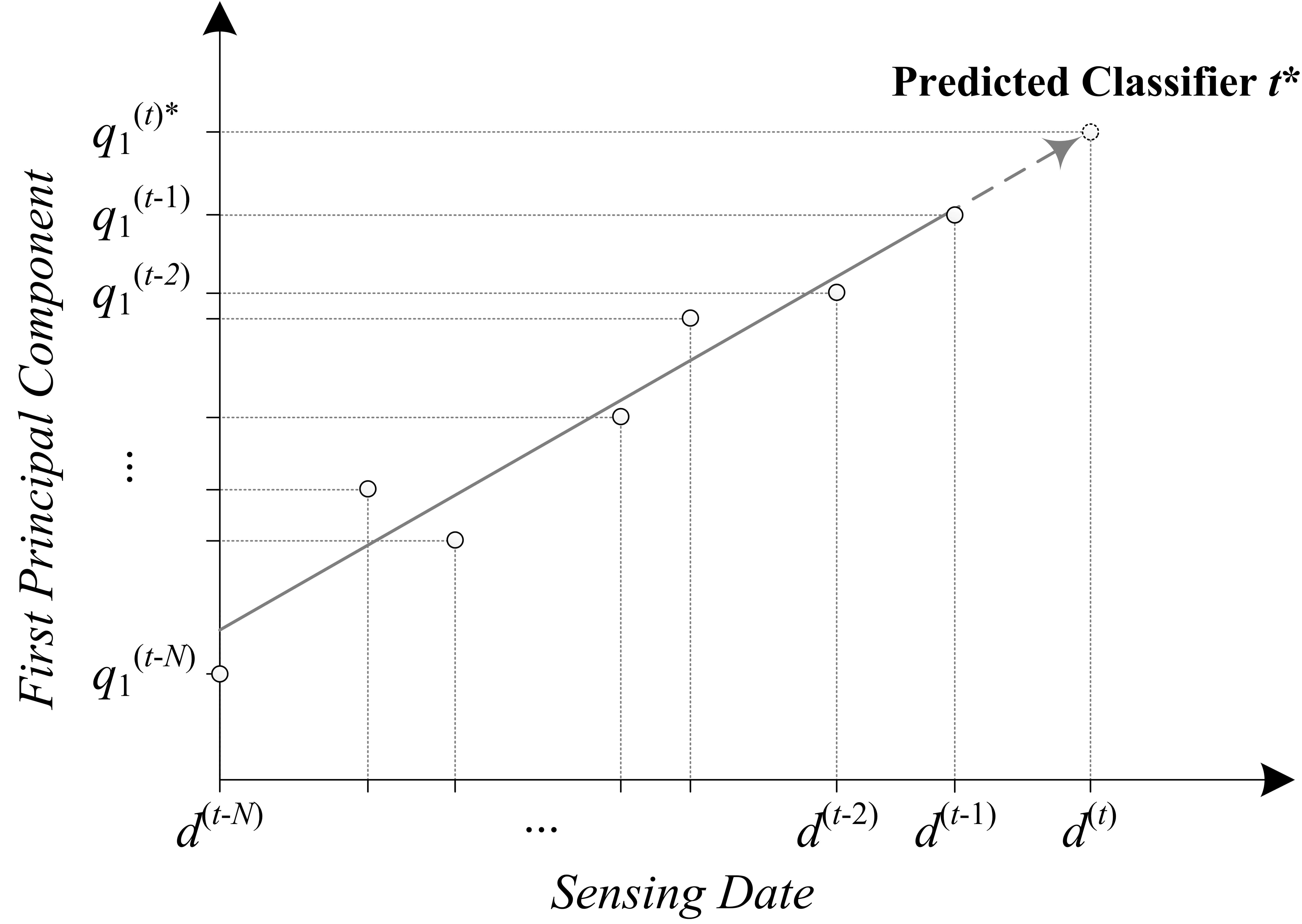}
\caption{Regression between first principal components and sensing dates.}\label{Fig: correlation}
\end{figure}

The next step is to map the maximum temporal dynamics direction of previous classifiers on a time scale. As shown in Fig. \ref{Fig: correlation}, the first principal components after the transformation, $\bar{q}_{1}^{\left( t-1 \right)}\text{, }\bar{q}_{1}^{\left( t-2 \right)}\text{, }\cdots \text{, }\bar{q}_{1}^{\left( t-N \right)}$, are now associated with the image sensing dates, $d^{(t-1)}$, $d^{(t-2)}$, $\cdots$, $d^{(t-N)}$, to show the intervals between each image collection, which can be irregular.
A regression is then performed in order to predict a classifier for Image $t$. The polynomial fitting function is selected in this study:
\begin{equation}
\begin{split}
\bar{q}_{1}^{\left( i \right)}={{a}_{0}}+{{a}_{1}}{{d}^{\left( i \right)}}+\cdots +&{{a}_{r}}{{\left( {{d}^{\left( i \right)}} \right)}^{r}}\text{,}\\
&i=t-1,\text{ }t-2\text{, }\cdots ,t-N,\\
\end{split}
\end{equation}
where $r$ is the order of the function, and ${a}_{0}$, ${a}_{1}$, $\cdots$, ${a}_{r}$ are the fitting coefficients. The sensing dates used here are integer counts of days starting from a consistent reference date.

It is worth noting that the order of the fitting function is selected according to the pattern of the temporal trend of classifiers. In order to achieve better fitting accuracies, higher-order polynomials may be adopted, but the overfitting risk will increase as well. So the balance between fitting accuracy and overfitting problem should be considered when choosing the fitting function.

After determining the fitting coefficients ${a}_{0}$ and ${a}_{1}$ with the data of previous classifiers, the first principal component of the predicted classifier for Image $t$, ${{\bar{q}}_1^{\left( t \right)*}}$, can be calculated as:
\begin{equation}
{{\bar{q}}_1^{\left( t \right)*}}={{a}_{0}}.
\end{equation}
The rest of the principal components $\bar{q}_{2}^{\left( t \right)*}$, $\bar{q}_{3}^{\left( t \right)*}$, $\cdots$, $\bar{q}_{m+1}^{\left( t \right)*}$ can be set to zero.

The inverse of the principal component transform is then performed, followed by the mean-decentralization (i.e., reverse of the mean-centralization), to obtain the predicted classifier in the original classifier parameter space:
\begin{equation}
{{\mathbf{p}}^{\left( t \right)*}}={{\mathbf{G}}^{-1}}{{\mathbf{\bar{q}}}^{\left( t \right)*}}+\frac{1}{N}\sum\limits_{i=t-1}^{t-N}{{{\mathbf{p}}^{\left( i \right)}}}.
\end{equation}

\subsection{Classifier Fine-Tuning}\label{ssec: fine-tuning}
A domain adaptation algorithm is developed as the second stage of the proposed SCT-SVM. It fine-tunes the predicted classifier ${{\mathbf{p}}^{\left( t \right)*}}$ to a more accurate position ${{\mathbf{p}}^{\left( t \right)}}$ with training samples of the current image (Image $t$). The algorithm is designed with anticipation that the true classifier should be located not far from the predicted position. This is a reasonable expectation, given that the predicted classifier is based on the historical information extracted from previous images and the changes are often gradual over time. The aim of the fine-tuning is set as to achieve a higher classification accuracy on the training samples of Image $t$. In this way, contributions from the predicted classifier and the training samples are incorporated and balanced in the algorithm, as detailed in the following.

The \emph{a priori} information provided by the predicted classifier is utilized by restricting the fine-tuned classifier from departing too far away from the predicted position. To achieve this goal, the distance between the first $m$ elements of
${{\mathbf{p}}^{\left( t \right)}}$ (i.e., ${{\mathbf{w}}^{\left( t \right)}}$) and ${{\mathbf{p}}^{\left( t \right)*}}$ (i.e., ${{\mathbf{w}}^{\left( t \right)*}}$) is minimized \cite{guo2017domain}.
More specifically, a set of non-negative slack variables $\boldsymbol{\mu }=\{{{\mu }_{j}}\}_{j=1}^{m}$ is introduced, where the $j$th element restricts the upper and lower bounds of the deviation of $p_{j}^{(t)}$ from $p_{j}^{(t)*}$, resulting in the following constraints:
\begin{equation}\label{Eq: constraints_1}
-{{\mu }_{j}}\le p_{j}^{\left( t \right)}-p_{j}^{\left( t \right)*}\le {{\mu }_{j}}\text{ and }{{\mu }_{j}}\ge 0\text{,}\quad j=1\text{, }2\text{, }\cdots \text{, }m.
\end{equation}

The information provided by the training samples of the current image (Image $t$) is utilized by taking into account the classification error of the fine-tuned classifier on the training samples $\left\{ {{\mathbf{x}}_{i}},{{y}_{i}} \right\}_{i=1}^{n}$, where ${{\mathbf{x}}_{i}}$ is the feature vector of the $i$th training sample and ${{y}_{i}}\in \left\{ +1\text{, }-1 \right\}$ is its corresponding label. Similar to those in the standard support vector machines, a set of non-negative slack variables $\boldsymbol{\xi }=\left\{ {{\xi }_{i}} \right\}_{i=1}^{n}$ are used to allow \emph{soft-margin} classification in order to tackle overlapping classes. Consequently, the following constraints are designed:
\begin{equation}\label{Eq: constraints_2_expanded}
\begin{split}
{{y}_{i}}\left( \sum\limits_{j=1}^{m}{p_{j}^{\left( t \right)}{{x}_{i,j}}+{{p}_{m+1}^{\left( t \right)}}} \right)\ge 1-{{\xi }_{i}}\text{ and }{{\xi }_{i}}\ge 0\text{,} & \\
i=1\text{, }2\text{, }\cdots \text{, }n\text{,} & \\
\end{split}
\end{equation}
where ${{x}_{i,j}}$ is the $j$th entry in ${{\mathbf{x}}_{i}}$.

With the constraints in Eqs. (\ref{Eq: constraints_1}) and (\ref{Eq: constraints_2_expanded}), the following optimization problem can be constructed:
\begin{equation}\label{Eq: optimization_problem}
\begin{split}
\text{min}\quad & \frac{1}{2}\sum\limits_{j=1}^{m}{{{\left( p_{j}^{(t)} \right)}^{2}}}+C\sum\limits_{i=1}^{n}{{{\xi }_{i}}}+F\sum\limits_{j=1}^{m}{{{\mu }_{j}}}\\
\text{s.t.}\quad & {{y}_{i}}\!\!\left(\! \sum\limits_{j=1}^{m}{p_{j}^{(t)}\!{{x}_{i,j}}\!+\!{{p}_{m+1}^{(t)}}} \!\!\right)\!\!\ge\!\! 1-{{\xi }_{i}}\text{ and }{{\xi }_{i}}\!\ge\! 0,\; i\!\!=\!\!1\text{,}2\text{,}\!\cdots\! \text{,}n, \\
& -{{\mu }_{j}}\le p_{j}^{(t)}-p_{j}^{(t)*}\le {{\mu }_{j}}\ \text{and}\ {{\mu }_{j}}\ge 0,\; j=1\text{,}2\text{,}\cdots \text{,}m. \\
\end{split}
\end{equation}\\
The first term in the objective function accounts for the margin space of the fine-tuned classifier. Like that in the standard support vector machines, the term aims to adjust the separating hyperplane to a position that generates the maximum margin space between classes. The second and third terms aim to minimize the degree of violation of the constraints. The positive constants $C$ and $F$ are regularisation parameters, which control the weights of the second and third terms relative to the first term. The values of $C$ and $F$ control the amounts of contributions from training samples and previous images, respectively. It is worth noting that the parameters, $F$ and $C$, need to be preset for the fine-tuning algorithm. The best combination of $F$ and $C$ can be determined through cross validation with grid search.

The dual form of the optimization problem in Eq. (\ref{Eq: optimization_problem}) can be obtained by constructing the following Lagrangian function $L$:
\begin{equation}\label{Eq: Lagrangian_prime}
\begin{split}
L=& \frac{1}{2}\sum\limits_{j=1}^{m}{{{\left( p_{j}^{(t)} \right)}^{2}}+C\sum\limits_{i=1}^{n}{{{\xi }_{i}}}}+F\sum\limits_{j=1}^{m}{{{\mu }_{j}}} \\
& -\sum\limits_{i=1}^{n}{{{\alpha }_{i}}\left[ {{y}_{i}}\left( \sum\limits_{j=1}^{m}{p_{j}^{(t)}{{x}_{i,j}}}+{{p}_{m+1}^{(t)}} \right)-\left( 1-{{\xi }_{i}} \right) \right]} \\
& -\sum\limits_{i=1}^{n}{{{\beta }_{i}}{{\xi }_{i}}}-\sum\limits_{j=1}^{m}{{{\gamma }_{j}}\left[ {{\mu }_{j}}-\left( p_{j}^{(t)}-p_{j}^{\left( t \right)*} \right) \right]} \\
& -\sum\limits_{j=1}^{m}{{{\delta }_{j}}\left[ \left( p_{j}^{(t)}-p_{j}^{\left( t \right)*} \right)+{{\mu }_{j}} \right]}-\sum\limits_{j=1}^{m}{{{\varepsilon }_{j}}{{\mu }_{j}}}, \\
\end{split}
\end{equation}
where $\left\{ {{\alpha }_{i}} \right\}_{i=1}^{n}$, $\left\{ {{\beta }_{i}} \right\}_{i=1}^{n}$, $\left\{ {{\gamma }_{j}} \right\}_{j=1}^{m}$, $\left\{ {{\delta }_{j}} \right\}_{j=1}^{m}$, and $\left\{ {{\varepsilon }_{j}} \right\}_{j=1}^{m}$ are non-negative Lagrange multipliers. Equating the partial derivatives of $L$ to zero with respect to $p_{j}^{(t)}$, ${{\xi }_{i}}$, and ${{\mu }_{j}}$ results in the following equations:
\begin{equation}\label{Eq: kkt_1}
p_{j}^{(t)}-\sum\limits_{i=1}^{n}{{{\alpha }_{i}}{{y}_{i}}{{x}_{i,j}}+{{\gamma }_{j}}-{{\delta }_{j}}}=0, \quad j=1\text{, }2\text{, }\cdots \text{, }m,
\end{equation}
\begin{equation}\label{Eq: kkt_2}
\sum\limits_{i=1}^{n}{{{\alpha }_{i}}{{y}_{i}}}=0,
\end{equation}
\begin{equation}\label{Eq: kkt_3}
{C}-{{\alpha }_{i}}-{{\beta }_{i}}=0,\quad i=1\text{, }2\text{, }\cdots \text{, }n,
\end{equation}
\begin{equation}\label{Eq: kkt_4}
F-{{\gamma }_{j}}-{{\delta }_{j}}-{{\varepsilon }_{j}}=0,\quad j=1\text{, }2\text{, }\cdots \text{, }m.
\end{equation}

Eliminating $p_{j}^{(t)}$, $C$, and $F$ by substituting Eqs. (\ref{Eq: kkt_1}--\ref{Eq: kkt_4}) into Eq. (\ref{Eq: Lagrangian_prime}) results in the following Lagrangian dual function $L_d$:
\begin{equation}\label{Eq: Lagrangian_dual}
\begin{split}
{{L}_{\mathrm{d}}}=& -\frac{1}{2}\sum\limits_{i=1}^{n}{\sum\limits_{k=1}^{n}{\sum\limits_{j=1}^{m}{{{\alpha }_{i}}{{\alpha }_{k}}{{y}_{i}}{{y}_{k}}{{x}_{i,j}}{{x}_{k,j}}}}}-\frac{1}{2}\sum\limits_{j=1}^{m}{{{\gamma }_{j}}^{2}} \\
& -\frac{1}{2}\sum\limits_{j=1}^{m}{{{\delta }_{j}}^{2}}+\sum\limits_{i=1}^{n}{\sum\limits_{j=1}^{m}{{{\gamma }_{j}}{{\alpha }_{i}}{{y}_{i}}{{x}_{i,j}}}} \\
& -\sum\limits_{i=1}^{n}{\sum\limits_{j=1}^{m}{{{\delta }_{j}}{{\alpha }_{i}}{{y}_{i}}{{x}_{i,j}}}}+\sum\limits_{j=1}^{m}{{{\gamma }_{j}}{{\delta }_{j}}}+\sum\limits_{i=1}^{n}{{{\alpha }_{i}}} \\
& -\sum\limits_{j=1}^{m}{p_{j}^{(t)*}\left( {{\gamma }_{j}}-{{\delta }_{j}} \right)}. \\
\end{split}
\end{equation}

Then the following dual optimization problem can be constructed, which is equivalent to the primal problem in Eq. (\ref{Eq: optimization_problem}) but has a simpler form:
\begin{small}
\begin{equation}\label{Eq: optimization_problem_dual}
\begin{split}
\underset{{{\alpha }_{i}},{{\gamma }_{j}},{{\delta }_{j}}}{\mathop{\mathrm{min}}}\,\quad & -{{L}_{d}}\quad  \\
\mathrm{s.t.}\quad & 0\le {{\alpha }_{i}}\le C,\quad i=1\text{, }2\text{, }\cdots \text{, }n, \\
& {{\gamma }_{j}}+{{\delta }_{j}}\le F,\; {{\gamma }_{j}}\ge 0,\; \mathrm{and}\; {{\delta }_{j}}\ge 0,\quad j=1\text{, }2\text{, }\cdots \text{, }m, \\
& \sum\limits_{i=1}^{n}{{{\alpha }_{i}}{{y}_{i}}}=0. \\
\end{split}
\end{equation}
\end{small}\\
This is a quadratic programming problem and can be solved with standard quadratic programming algorithms \cite{nocedal2006numerical}. After determining ${{\alpha }_{i}}$, ${{\gamma }_{i}}$, and ${{\delta }_{i}}$, the parameters for the fine-tuned classifier ${{\mathbf{p}}^{\left( t \right)}}$ can be calculated as:
\begin{equation}\label{Eq: w_final}
p_{j}^{(t)}=\sum\limits_{i=1}^{n}{{\alpha }_{i}}{{y}_{i}}{{x}_{i,j}}-{{\gamma }_{j}}+{{\delta }_{j}},\quad j=1\text{, }2\text{, }\cdots \text{, }m,
\end{equation}
\begin{equation}\label{Eq: b_final}
{{p}_{m+1}^{(t)}}=\frac{1}{{{N}_\mathcal{S}}}\sum\limits_{i\in \mathcal{S}}{\left[ \frac{1}{{{y}_{i}}}-\sum\limits_{j=1}^{m}p_{j}^{\left( t \right)}{{x}_{i,j}} \right]},
\end{equation}
where $\mathcal{S}$ and ${N}_\mathcal{S}$ are the index set and the total number of support vectors, respectively.

\subsection{Discussion}
\subsubsection{Advantages of Classifier-Level Approach}

\quad

The proposed SCT-SVM is a classifier-level approach. It transfers the classifier-level knowledge, i.e. the classifier parameters of a set of previous classifiers, to the incoming image. Compared with the data-level approaches that require the availability of pixels and/or their corresponding labels (raw data) from previous images, such as the Domain Adaptation SVM (DA-SVM) \cite{bruzzone2010domain} and the Domain Transfer SVM (DT-SVM) \cite{duan2009domain}, the proposed approach has the following two merits. Firstly, the required computational load can be reduced considering that the size of raw data is usually large. Secondly, the proposed algorithm is applicable even if the raw data of previous images are inaccessible (e.g., private or no longer stored). Therefore, compared with data-level approaches, the proposed classifier-level approach is more efficient and more feasible.

\subsubsection{Comparing Proposed Fine-tuning Algorithm to State-of-the-Art Algorithms}

\quad

The second step of the proposed method is fine-tuning. There are two existing domain adaptation algorithms in the literature that are able to handle the same task: the Adaptive SVM (A-SVM) \cite{yang2007cross,yang2008learning} and the Projective Model Transfer SVM (PMT-SVM) \cite{aytar2011tabula}. The A-SVM enables the estimation of a fine-tuned target classifier by augmenting a perturbation term to the predicted classifier. However, it has been found that the A-SVM tends to generate a small margin space for the fine-tuned classifier \cite{aytar2011tabula}. The other algorithm, PMT-SVM, is able to estimate a fine-tuned classifier without penalizing its margin maximization. The limitation, however, is that it restricts the included angle between the predicted and fine-tuned classifiers to a value equivalent to or less than 90$^\circ$ in the feature space. This restriction leads to a decreased performance when the underlying optimal classifier is positioned at an angle greater than 90$^\circ$ from the predicted position.

The proposed fine-tuning algorithm considers a maximum-margin-space term (i.e., the first term of the objective function in Eq. (\ref{Eq: optimization_problem})), which can alleviate the small-margin-space problem that exists in the A-SVM. The proposed algorithm is also able to fine-tune a predicted classifier into a position larger than 90$^\circ$ as no restriction is applied on the fine-tuning angle, which is an improvement from the PMT-SVM. The proposed domain adaptation algorithm is named as TA-SVM for convenience, and it is compared with A-SVM and PMT-SVM in the experiments.

\section{Experiments and Results}\label{sec: experiment}

\subsection{Study Area and Data Sets}

\subsubsection{Study Area Description}

\quad

\begin{figure*}[!htb]
\centering
\includegraphics[width=6.5in]{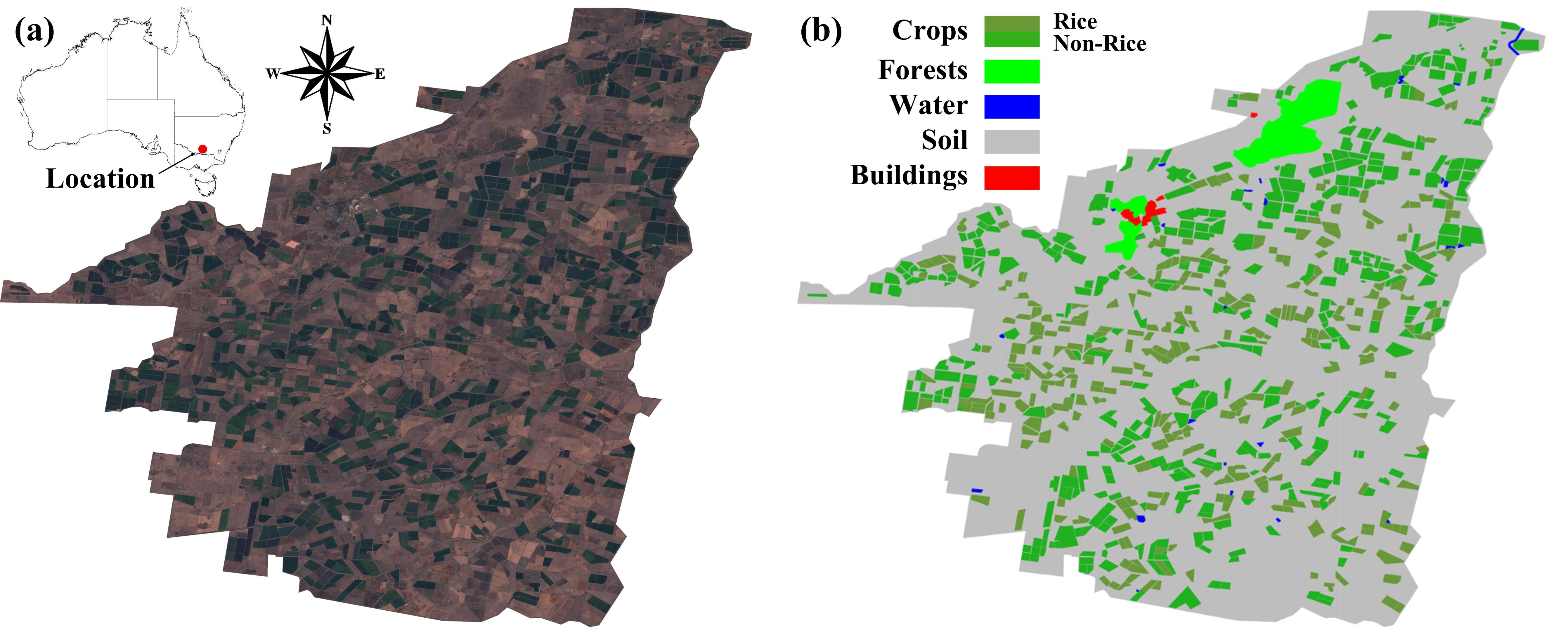}
\caption{(a) Location and color composite map of the study area (Sentinel-2A image on Mar. 09th, 2017; R: Band 4, G: Band 3, B: Band 2); (b) Ground reference map.}\label{Fig: location}
\end{figure*}

The experimental area is located at the Coleambally Irrigation Area in the southwest of New South Wales, Australia (145$^\circ$40'27''$–-$146$^\circ$08'12'' E, 34$^\circ$40'37''$–-$35$^\circ$03'28'' S), as shown in Fig. \ref{Fig: location}a. A high water-consuming crop, rice, is widely planted in this area. Its growing period lasts for about six months, roughly from October of the first year to April of the second year. Rice is the only crop in the area that needs to be flooded during its growing period \cite{xiao2005mapping}. This makes its spectral characteristics distinguishable from other summer crops. The area has a semiarid climate (K\"{o}ppen-Geiger BSk). The annual rainfall is about 400 mm, distributed evenly throughout the year. Due to the semi-arid climate and high water consumption of the crop industry, water supplement in this area largely depends on irrigation from the Murrumbidgee River. Therefore, timely and accurate mapping of land cover distributions is critical for the local Irrigation Company to make appropriate decisions on water budgeting and allocation, and to conduct environmental assessments. In this study, a set of multitemporal Sentinel images was firstly used to classify different land cover types in the area, for the 2016-2017 summer growing season. Then rice and non-rice crops were separated with a set of multitemporal Landsat images of the same season.

\subsubsection{Ground Reference Data}

\quad

There were five land cover classes identified in the study area: crops (including rice and non-rice), forests, water, soil, and buildings, as shown in the ground reference map in Fig. \ref{Fig: location}b. A total of 200 pixels were selected for each class, and labelled via photointerpretation with the help of the farmland boundary data by the Coleambally Irrigation Company. These labelled pixels were used for the multitemporal Sentinel experiment. The crops class was further divided into rice and non-rice subclasses. For each subclass, 200 pixels were selected and labelled based on the rice distribution map provided by the Sunrice Milling Company, and used for the multitemporal Landsat experiment.

\subsubsection{Multitemporal Sentinel Data}

\quad

Five cloud-free Sentinel-2A optical images were collected from the Google Earth Engine platform. The images were sensed on Jan. 28th, Feb. 07th, Feb. 27th, Mar. 09th, and Mar. 19th, respectively, in the year of 2017. Each image consisted of 13 spectral bands covering the visible, near-infrared, and short-wave-infrared spectral regions. The images were provided in top of atmosphere (TOA) reflectance values with radiometric and geometric corrections applied (Level-1C data). The bands with 10 m (Bands 2, 3, 4, and 8) and 20 m (Bands 5, 6, 7, 8a, 11, and 12) spatial resolutions were reduced down to 60 m resolutions in order to make them consistent with the other bands (Bands 1, 9, and 10) that had 60 m resolutions.

In the experiments, the first four images are used as previous images to assist the training and classification of the fifth image. Classifier prediction and fine-tuning results based on this data set are provided in the following Subsections \ref{ssec: classifier_prediction_results} and \ref{ssec: classifier_fine-tuning_results}, respectively. Classification results with the proposed approach are compared with those obtained without the assistance from previous images in Subsection \ref{ssec: classification_results}.

\subsubsection{Multitemporal Landsat Data}

\quad

The multitemporal Landsat dataset consisted of ten cloud-free Landsat-8 multispectral images collected from the Google Earth Engine platform. Their sensing dates were from Dec. 12th, 2016 to Mar. 27, 2017, covering the major growing period of rice. These images were provided in atmospherically corrected surface reflectance values with a spatial resolution of 30m. Among all the available bands, five visible and near-infrared bands (Bands 1, 2, 3, 4, and 5) and 2 short-wave-infrared bands (Bands 6 and 7) were used in this study. This dataset was used to investigate the sensitivity of the proposed method to previous classifiers' errors. The results are shown in Section \ref{Sec: error_analyses}.

\subsection{Classifier Prediction Results}\label{ssec: classifier_prediction_results}
This section presents the classifier prediction results obtained with the proposed SCT-SVM approach. The temporal movements of classifiers, which were driven by the variations of class data, were firstly analyzed. Then fitting functions were used to fit the movement patterns of classifiers. Classifiers predicted with different fitting functions were compared with the true classifiers to assess the prediction accuracies.

\subsubsection{Temporal Variations of Classifiers}

\quad

The correlation between spectral features and sensing dates was firstly analyzed. Bands 4 (red) and 8 (near-infrared) were selected for the analysis. Linear correlation coefficients were calculated for the five land cover classes, with the results shown in Tab. \ref{Tab: correlation}. The highest temporal correlations were found for crops. The coefficient for the near-infrared band was -0.8306, indicating a strong negative correlation with time. The red band showed a positive but weaker correlation with a coefficient of 0.5213. Another vegetative class, forests, showed similar but less significant temporal correlations with coefficients of 0.4209 and -0.4699 for the red and infrared bands, respectively. The spectral features of the three non-vegetative classes, water, soil, and buildings, showed insignificant temporal correlations, with all the coefficients lower than 0.1.

\begin{figure}[!htb]
\centering
\includegraphics[width=3.45in]{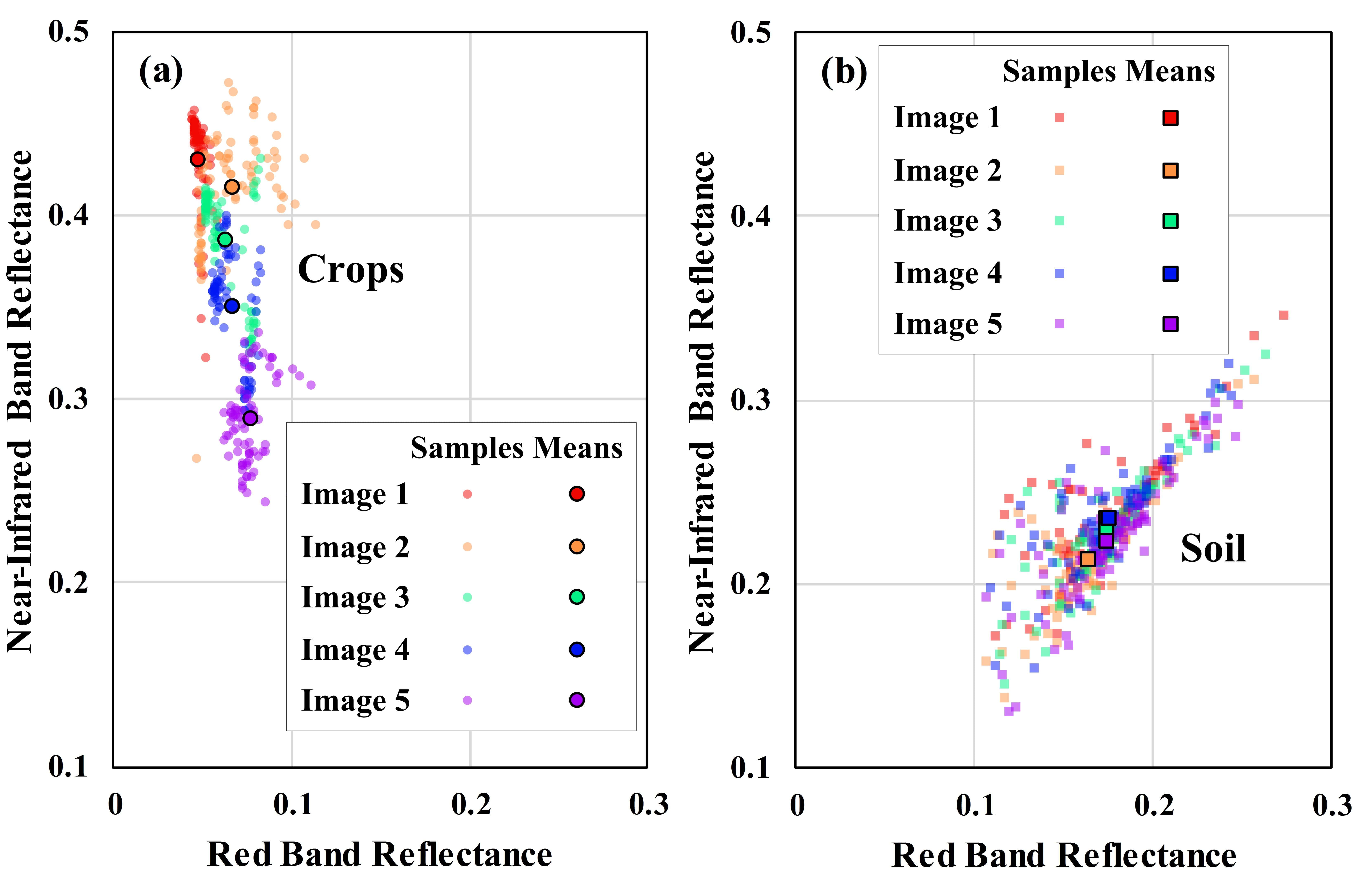}
\caption{Temporal variations of the class data of crops and soil.}\label{Fig: trend}
\end{figure}

\begin{figure}[!htb]
\centering
\includegraphics[width=3.45in]{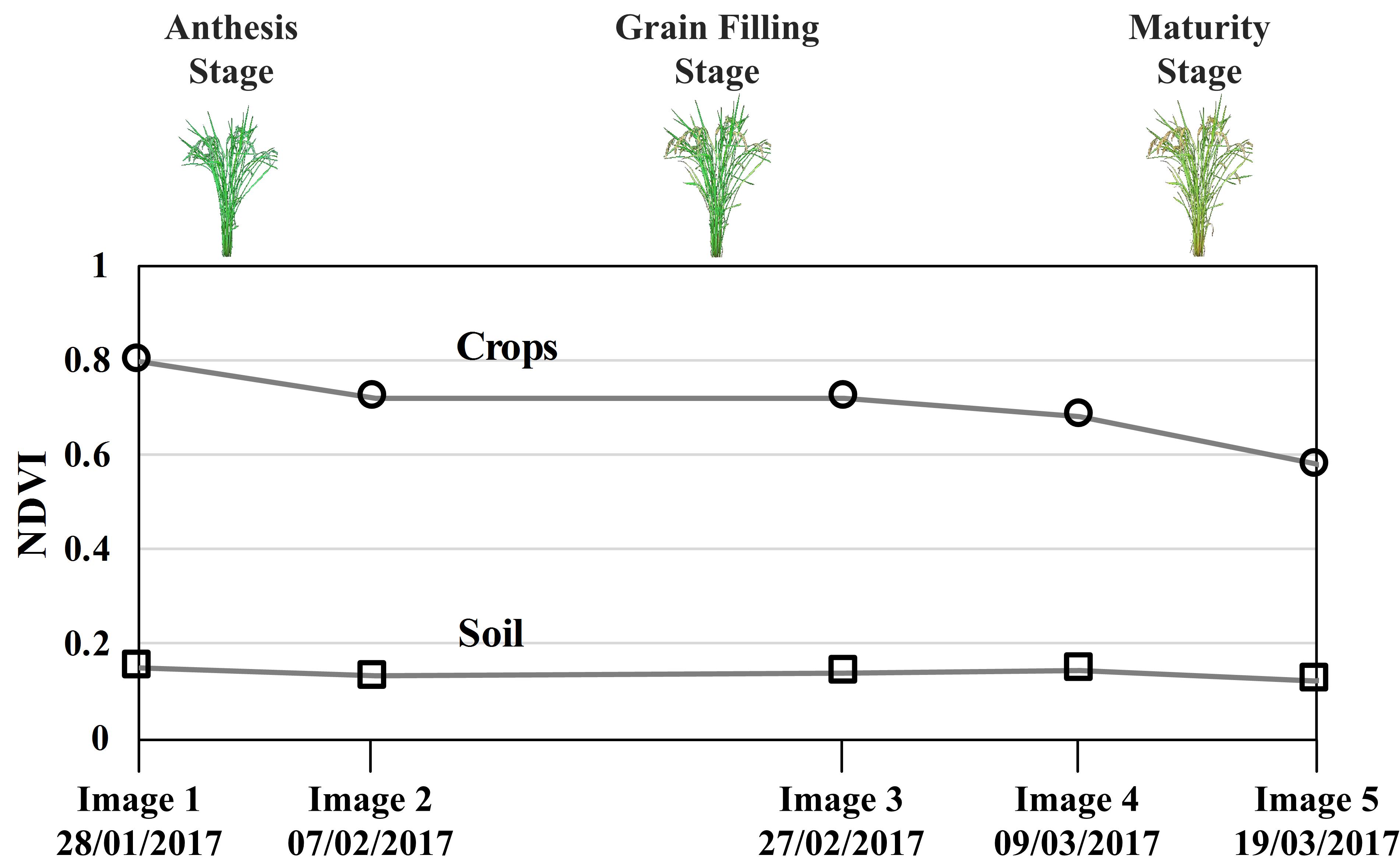}
\caption{Multitemporal Normalized Differential Vegetation Index (NDVI) of crops and soil and the corresponding growing stages of crops.}\label{Fig: NDVI}
\end{figure}

More detailed analysis of the temporal correlations for crops and soil is shown in Fig. \ref{Fig: trend}. By examining the class data of different images, it was found that the means of crops gradually moved from the top left towards the bottom right in the red/near-infrared feature space. In contrast, the means of soil remained relatively stable. Figure \ref{Fig: NDVI} explains the movement patterns of data of these classes from a phenological perspective. It was found that the five images covered the mid-late growing periods of crops. During these periods, the crop leaves turned from green to yellow, accompanied by a gradual decrease in the value of Normalized Differential Vegetation Index (NDVI). The NDVI of soil remained stable with time, suggesting an insignificant temporal variation.

\begin{table}
\caption{Linear correlation coefficients between spectral features and sensing dates.}\label{Tab: correlation}
  \begin{tabular}{p{1.0cm}p{1.0cm}p{1.0cm}p{1.0cm}p{1.0cm}p{1.2cm}}
    \hline
  &Crops &Forests &Water &Soil &Buildings\\ \hline
  Band 4 &0.5213 &0.4209 &-0.0832 &0.0745 &-0.0023\\
  Band 8 &-0.8306 &-0.4699 &0.0531 &0.0135 &0.0012\\ \hline
  \end{tabular}
\end{table}

\begin{figure}[!htb]
\centering
\includegraphics[width=2.8in]{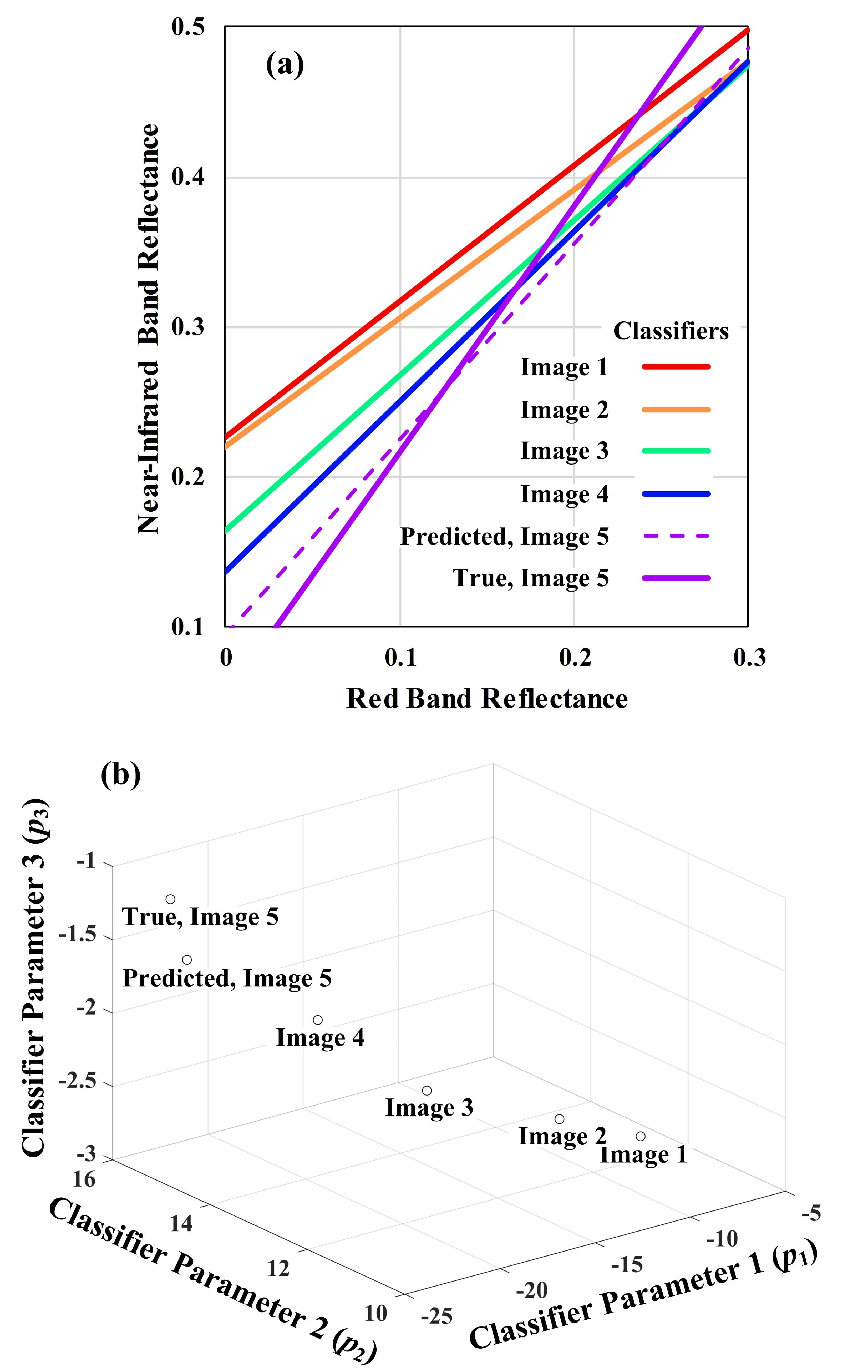}
\caption{Multitemporal positions of the crops-against-soil classifier in (a) the feature space and (b) the classifier parameter space.}\label{Fig: space}
\end{figure}

Driven by the temporal trends of class data, the position of the classifier separating crops and soil drifted with time as well. Figures \ref{Fig: space}(a) and \ref{Fig: space}(b) show the movement of the crops-against-soil classifier in the feature space and classifier parameter space, respectively. It was found that the classifier position moved consistently with time. Based on the first four classifier positions, a predicted position was estimated for the fifth classifier. Though the predicted classifier was not exactly located at the true position, it was a good \emph{a priori} guess that could provide advantageous help in seeking the true classifier.

\subsubsection{Prediction Accuracies with Different Orders of Fitting Functions}

\quad

Different orders of fitting functions were tested for classifier predictions. For each classifier separating two classes, the Euclidean distance between the predicted and true positions was calculated, in order to quantitatively measure the prediction accuracy. The results are shown in Tab. \ref{Tab: fitting}. The highest accuracy for each classifier is shown in bold typeface. It was found that, for classifiers that involved one or more vegetative classes (crops and/or forests), second-order polynomial fitting produced the best prediction results. For classifiers that separated two non-vegetative classes, first-order polynomial fitting was the best choice.

\begin{table}
\caption{Classifier prediction accuracies with different orders of polynomial fitting functions.}\label{Tab: fitting}
  \begin{tabular}{p{2.30cm}p{1.60cm}p{1.60cm}p{1.60cm}}
    \hline
  Classifier &First-Order &Second-Order &Third-Order\\ \hline
  Crops/Forests &3.068 &\textbf{1.729} &8.235\\
  Crops/Soil &2.515 &\textbf{1.324} &7.142\\
  Crops/Water &2.613 &\textbf{1.771} &6.851\\
  Crops/Buildings &1.791 &\textbf{1.271} &5.982\\
  Forests/Soil &2.902 &\textbf{2.314} &9.007\\
  Forests/Water &3.181 &\textbf{2.117} &8.967\\
  Forests/Buildings &3.452 &\textbf{2.209} &8.437\\
  Soil/Water &\textbf{0.912} &1.127 &5.426\\
  Soil/Buildings &\textbf{0.814} &1.477 &3.251\\
  Water/Buildings &\textbf{1.113} &1.674 &4.865\\ \hline
  \end{tabular}
\end{table}

\begin{figure}[!htb]
\centering
\includegraphics[width=3.4in]{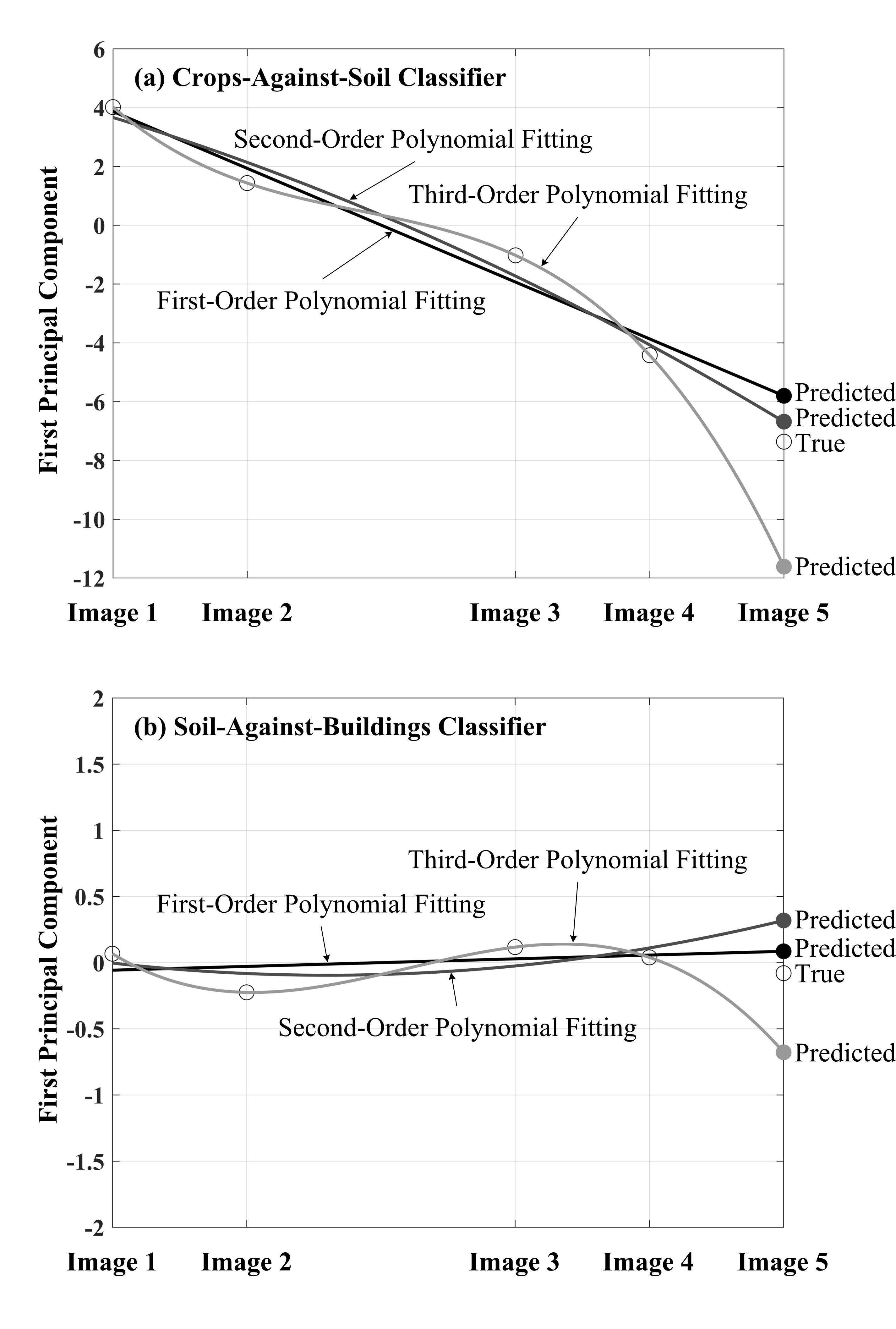}
\caption{Prediction accuracies for the crops-against-soil and soil-against-buildings classifiers with different fitting functions.}\label{Fig: polynomial}
\end{figure}

More detailed analyses of the prediction accuracies for the crops-against-soil and soil-against-buildings classifiers are shown in Fig. \ref{Fig: polynomial}. For the crops-against-soil classifier, compared with first-order polynomial, the second-order polynomial was more capable of describing the non-linear movement of classifier's positions. The third-order polynomial generated the worst result due to the overfitting effect. The non-linear movement was less obvious for the soil-against-buildings classifier, so the first-order fitting produced the best accuracy in this case.

\subsection{Classifier Fine-Tuning Results}\label{ssec: classifier_fine-tuning_results}
The proposed SCT-SVM approach includes a new domain adaptation algorithm, TA-SVM, to fine-tune classifiers from the predicted positions to more accurate positions. In this section, the performance of TA-SVM in classifier fine-tuning was firstly compared with two existing algorithms, A-SVM \cite{yang2007cross} and PMT-SVM \cite{aytar2011tabula}. Then the impacts of two TA-SVM parameters, $F$ and $C$, on fine-tuning accuracies were analyzed and discussed.

\subsubsection{Comparisons with A-SVM and PMT-SVM Algorithms}

\quad

The three fine-tuning algorithms, A-SVM \cite{yang2007cross}, PMT-SVM \cite{aytar2011tabula} and the proposed TA-SVM, were compared. In this analysis, the number of training samples for each class of the fifth image ($N_\mathcal{T}$) ranged from 5 to 50, in order to analyze and compare the algorithms under different levels of training data insufficiency. Classification accuracies achieved with the fine-tuned classifiers on the rest of the labelled samples were calculated as a measure to evaluate the three algorithms.

\begin{figure}[!htb]
\centering
\includegraphics[width=3.45in]{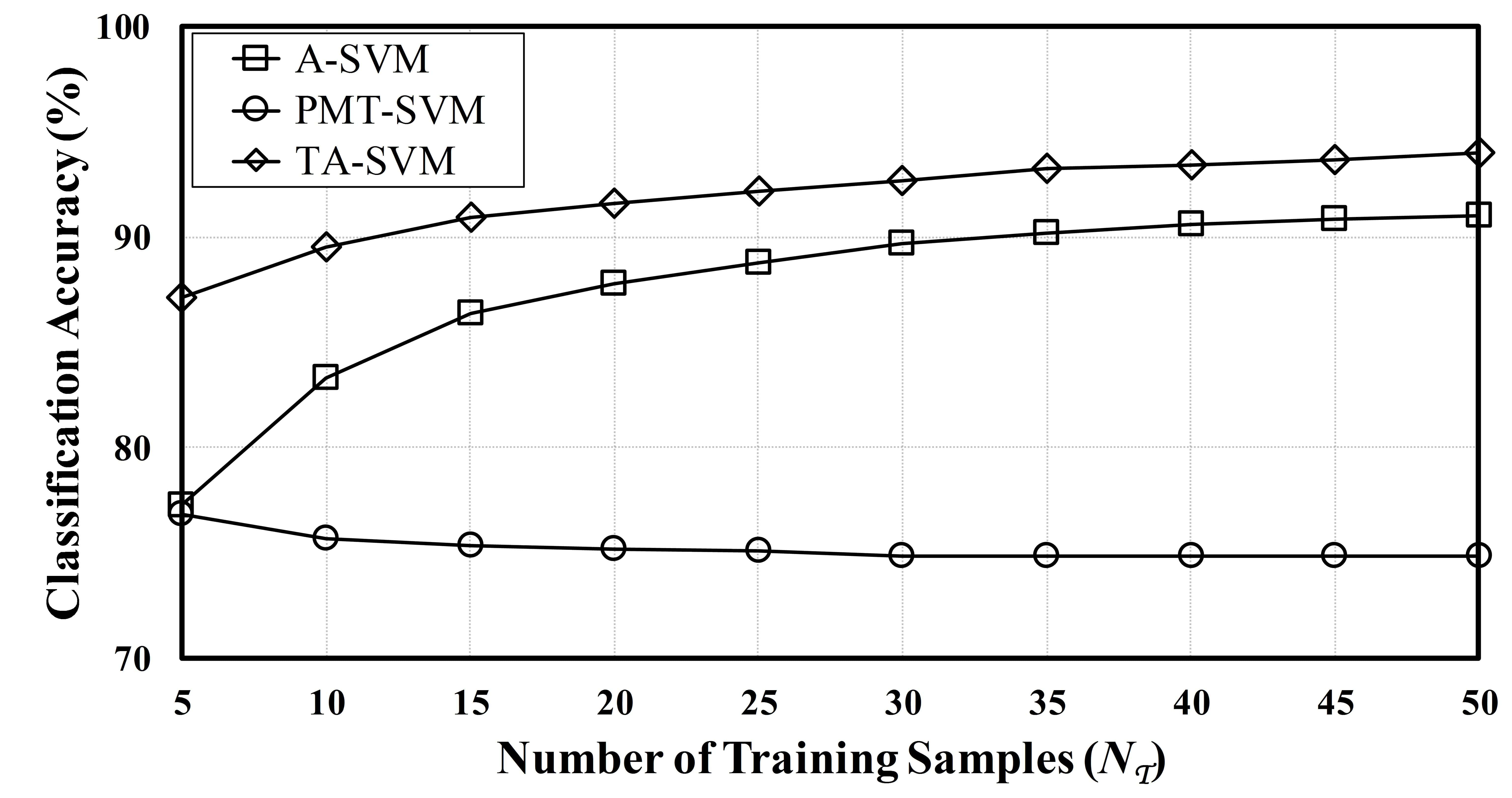}
\caption{Comparison of the three fine-tuning algorithms, A-SVM, PMT-SVM, and the proposed TA-SVM, with different Number of Training Samples from the current image.}\label{Fig: threeAlgorithms}
\end{figure}

The results are shown in Fig. \ref{Fig: threeAlgorithms}. It was found that the accuracies of A-SVM and TA-SVM increased with $N_\mathcal{T}$, but the TA-SVM generated relatively higher accuracies especially when training data were insufficient ($N_\mathcal{T}$\textless10). The other algorithm, PMT-SVM, produced low accuracies for all $N_\mathcal{T}$ levels. The results indicated that the TA-SVM outperformed the other two algorithms in classifier fine-tuning.

\begin{figure}[!htb]
\centering
\includegraphics[width=2.5in]{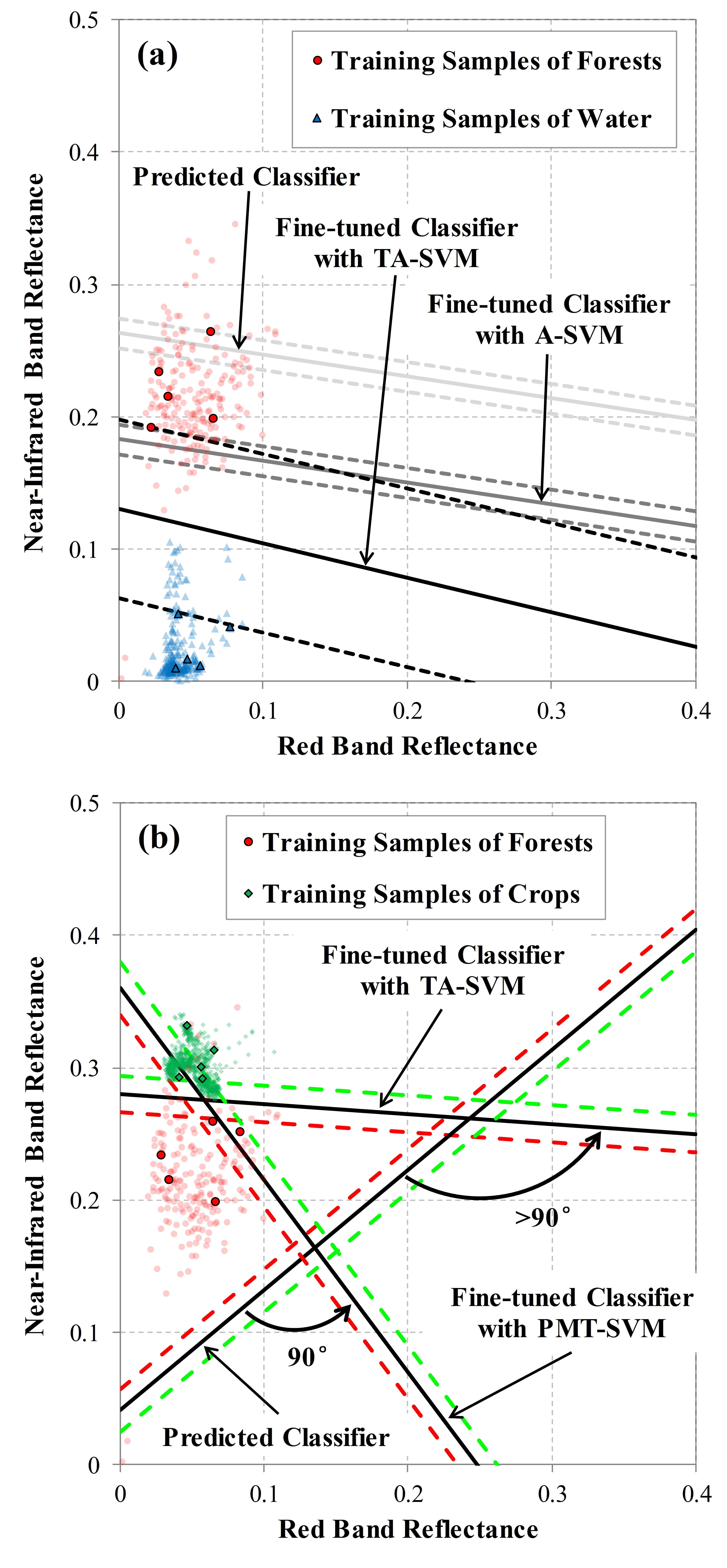}
\caption{Comparison of the behaviors of A-SVM, PMT-SVM, and TA-SVM for the fine-tuning of (a) the forests-against-water classifier  and (b) the crops-against-forests classifier.}\label{Fig: state}
\end{figure}

In-depth comparisons of the three algorithms are given in Fig. \ref{Fig: state}, where the behaviours of these algorithms were analyzed with two classifier fine-tuning cases. In the first case shown in Fig. \ref{Fig: state}(a), it was found that the classifier fine-tuned with A-SVM generated a relatively small margin space. This made the classifier be located at a biased position and produce a low classification accuracy. Differently, the classifier fine-tuned with TA-SVM was positioned at the maximum-margin location where a higher classification accuracy was achieved. In the second case shown in Fig. \ref{Fig: state}(b), the classifier fine-tuned with PMT-SVM was restricted to 90$^\circ$ with the predicted classifier, while the classifier fine-tuned with the proposed TA-SVM was able to be positioned at an obtuse angle where higher classification accuracy could be achieved.

\subsubsection{Impacts of Regularization Parameters}

\quad

\begin{figure}[!htb]
\centering
\includegraphics[width=3.4in]{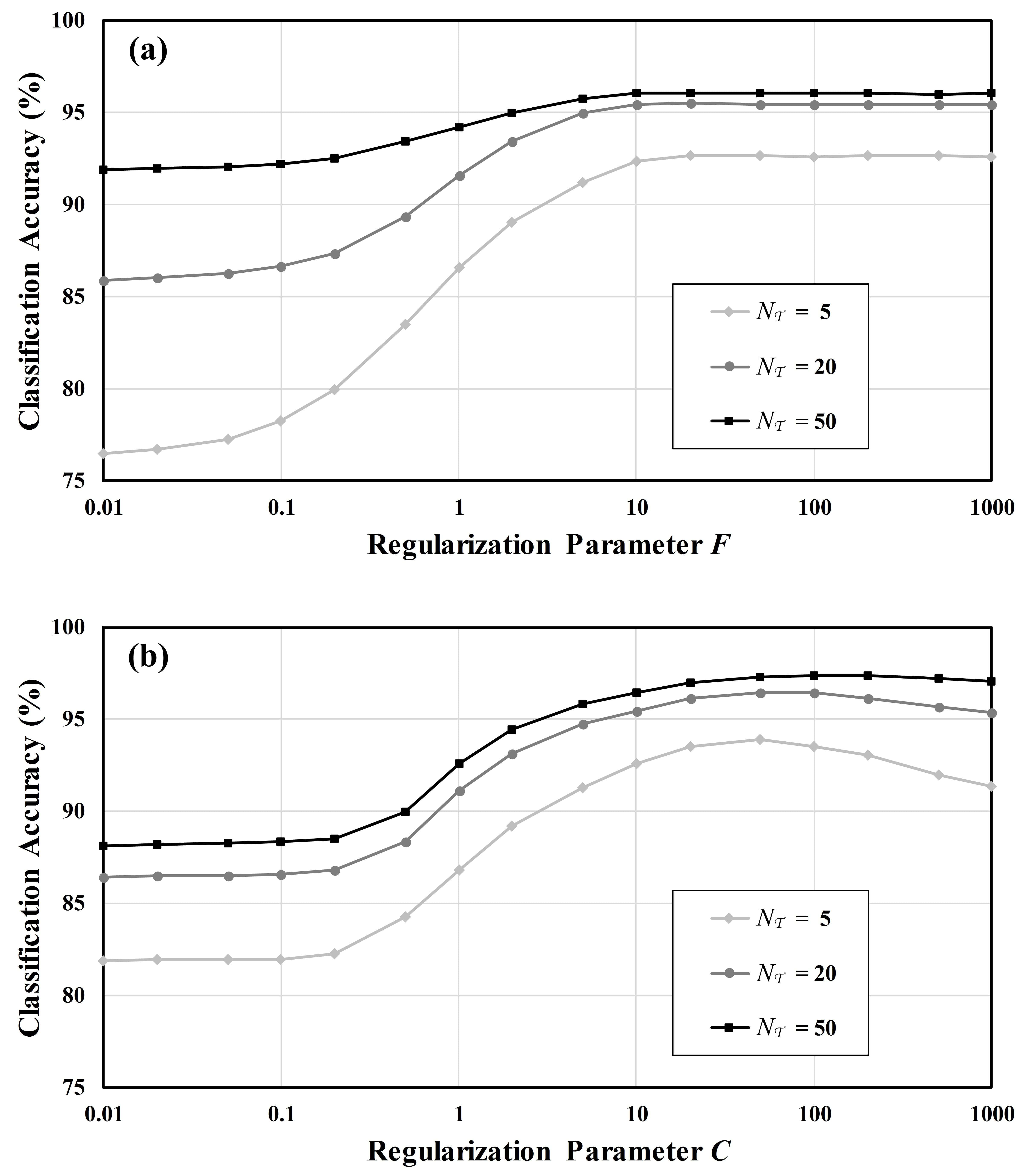}
\caption{Impact of the regularization parameters $F$ and $C$ on overall classification accuracy with different number of training samples for each class of the fifth image ($N_\mathcal{T}$).}\label{Fig: fandc}
\end{figure}

The two regularization parameters adopted in TA-SVM, $F$ and $C$, control the contributions of information from previous images and current training samples, respectively. Here experiments were conducted to analyze the impacts of these parameters on the fine-tuning results. In the experiments, the number of training samples for each class of the fifth image ($N_\mathcal{T}$) was set to 5, 20, and 50, corresponding to training sample cases that were insufficient, moderate and sufficient, respectively. The rest of the labelled samples were used to assess the accuracies of the classifiers fine-tuned with different settings of regularization parameters. The results are presented in Fig. \ref{Fig: fandc}.

Fig. \ref{Fig: fandc}a shows the results for $F$. It was found that, if training samples were limited ($N_\mathcal{T}$ = 5), higher classification accuracy was achieved with a larger $F$ value. Low classification accuracy was observed (76.37$\%$) when $F$ was set to a small value of 0.01, as the information from previous images played an insignificant role. When $F$ was set to 10 or higher, it was able to provide enough weight for the information from previous images, and considerably increased classification accuracies were achieved (\textgreater 92$\%$). The importance of $F$ was less critical if training samples were sufficient ($N_\mathcal{T}$ = 50). When $F$ was increased from 0.01 to 10, the increase in classification accuracy was from 91.87$\%$ to 95.98$\%$. This increase was less significant compared with the training sample insufficiency case (i.e., $N_\mathcal{T}$ = 5).

Fig. \ref{Fig: fandc}b shows the results for $C$. When $C$ was set to a small value such as 0.01, the contribution of information from training samples was low. In this case, low classification accuracies were observed for all the three $N_\mathcal{T}$ levels. With increased $C$ values, higher classification accuracies were achieved. However, decreases in classification accuracy occurred when $C$ was larger than 100, because the classifier training relied too heavily on the insufficient training samples, and the contribution from previous images was suppressed. This phenomenon was more obvious for a lower $N_\mathcal{T}$.

These results indicate that, in order to achieve satisfactory classification accuracy, it is necessary to adequately incorporate and balance the contributions from previous images and training samples by appropriately setting $F$ and $C$.

\subsection{Classification Results with and without Assistance from Previous Images}\label{ssec: classification_results}
We used the proposed SCT-SVM approach to classify the fifth image. The \emph{one-against-one} strategy was used to extend the binary classifiers to multiple classes. According to the results presented above, regularization parameters $F$ and $C$ were set to 20 and 50, respectively. The second-order polynomial fitting function was selected for classifiers that involved one or more vegetative classes, while for the rest of the classifiers the first-order polynomial was selected. With the assistance of the first four images, the fifth incoming image was trained and classified under the training sample insufficiency and sufficiency situations ($N_\mathcal{T}$ = 5 and 50). The training and classification were repeated 10 times with different sets of training samples selected from the labelled data. The classification accuracies were compared with those obtained when SVM was applied directly to the fifth image without using the \emph{a priori} knowledge from previous images (Dir-SVM).

\begin{figure}[!htb]
\centering
\includegraphics[width=3.4in]{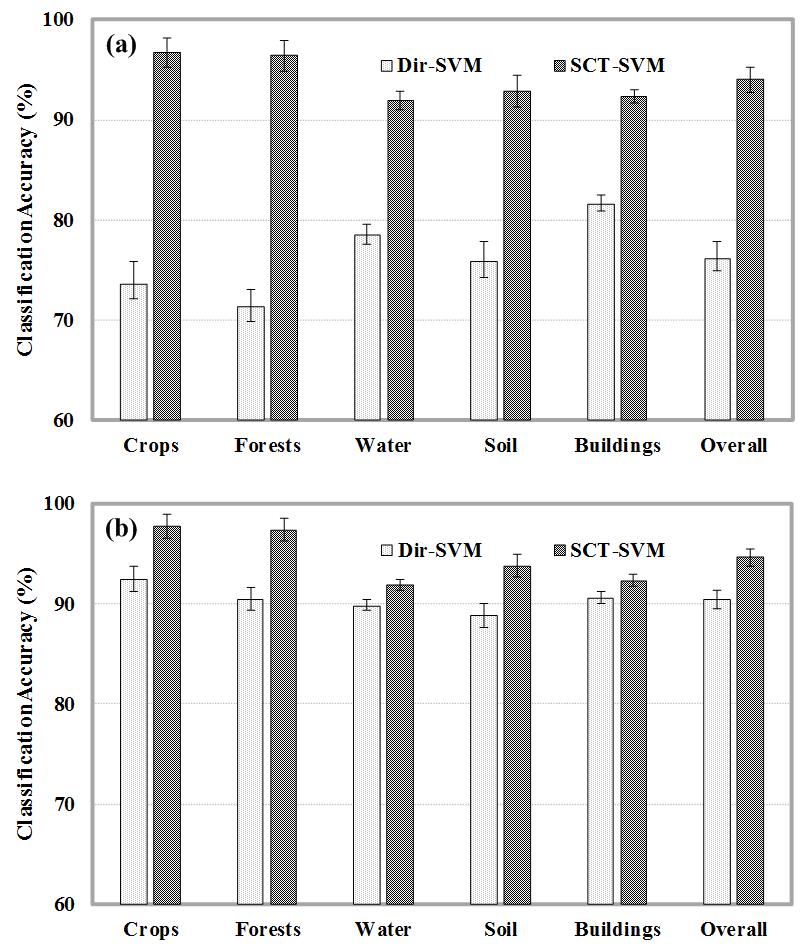}
\caption{Comparison of classification accuracies with and without the \emph{a priori} knowledge from previous images under (a) the training sample insufficiency situation ($N_\mathcal{T}$ = 5) and (b) the training sample sufficiency situation ($N_\mathcal{T}$ = 50). Error bars indicate the standard deviations of classification accuracies obtained with different sets of training samples.}\label{Fig: classification}
\end{figure}

The results are shown in Fig. \ref{Fig: classification}. It was found that, when training data were insufficient, the overall classification accuracy was improved from 76.18$\%$ to 94.02$\%$ if the \emph{a priori} knowledge provided by previous images was utilized, as shown in Fig. \ref{Fig: classification}a. The greatest improvements were found for crops and forests, two vegetative land cover types that share similar spectral characteristics. The remaining three land cover types showed improved classification accuracies as well, though the improvements were comparatively lower. These results demonstrate that, when the training samples are limited, the utilization of \emph{a priori} knowledge from previous images is advantageous to enhance classification accuracy. For the training sample sufficiency case shown in Fig. \ref{Fig: classification}b, the proposed SCT-SVM approach still generated better classification results than Dir-SVM, though the improvements were smaller compared with the small training sample case (Fig. \ref{Fig: classification}a). This indicates that, when training data of the current image are sufficient, \emph{a priori} knowledge from previous images is less critical.

\subsection{Error Analysis}\label{Sec: error_analyses}

In this Section, the robustness of the proposed method to previous classifiers' errors was analyzed with a multitemporal Landsat dataset in the Coleambally area. Rice and non-rice were selected as the classes of interest. In the error analysis, poor classifiers were simulated by adding a few levels of noises to the underlying optimal classifiers parameters. According to the classification accuracies they offered, these poor classifiers were divided into three groups: 60$\%$$\sim$70$\%$, 70$\%$$\sim$80$\%$, and 80$\%$$\sim$$A_{max}$ (i.e., the maximum accuracies achieved by the underlying best classifiers), with each group consisting of 100 classifiers. The following two experiments were conducted based on these simulated classifiers.

\subsubsection{Responses of Classifier Prediction and Fine-tuning to Previous Classifiers' Errors}

\quad

\begin{figure}[!htb]
\centering
\includegraphics[width=3.2in]{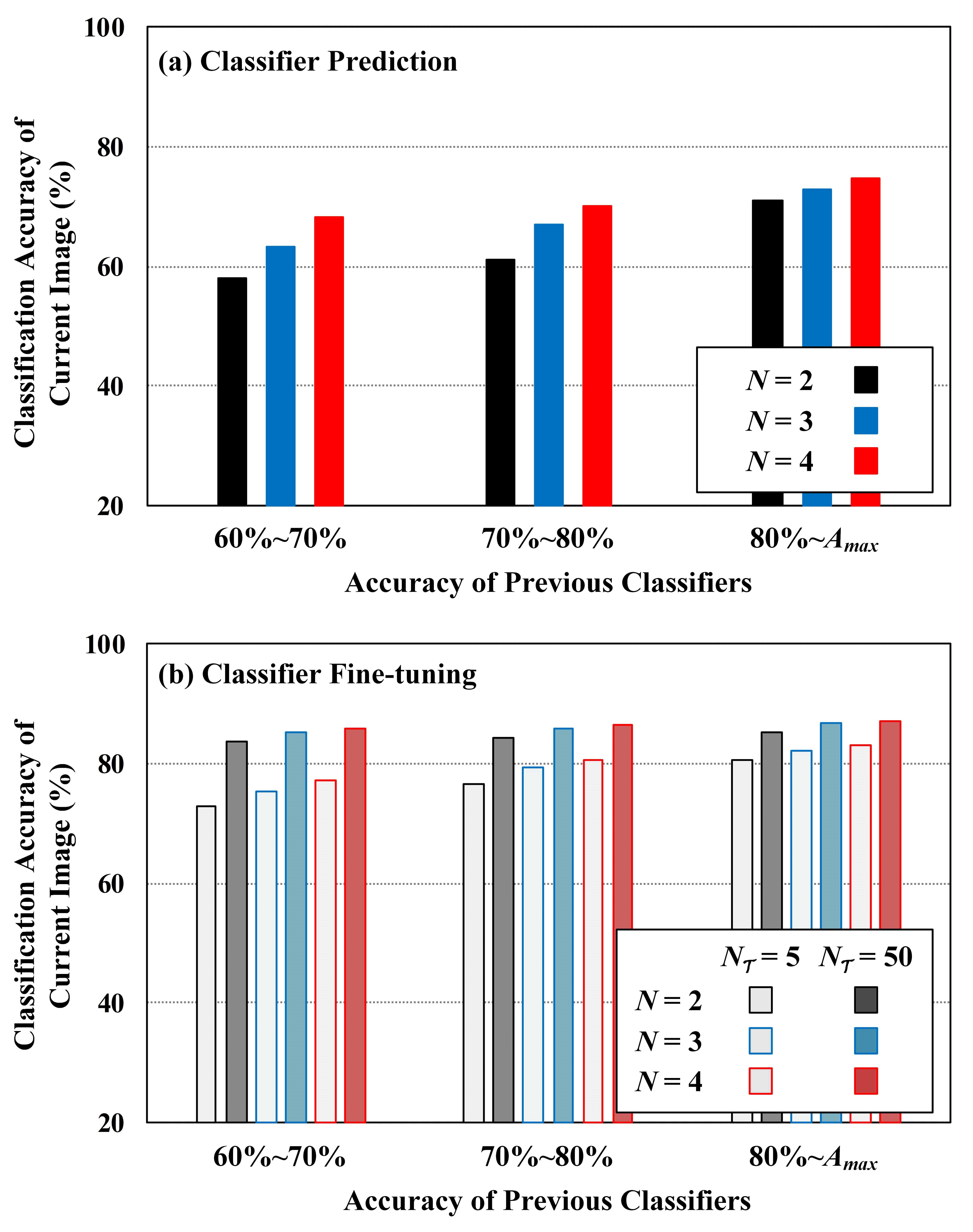}
\caption{Classifier prediction and fine-tuning accuracies. $N$ is the number of previous classifiers used. $N_\mathcal{T}$ is the number of training samples from the current image used.}\label{Fig: error_n1}
\end{figure}

In this experiment, classifier prediction and fine-tuning accuracies using different levels of previous classifiers' accuracies were investigated. The influences of another two factors, the number of previous classifiers ($N$) and the number of training samples of the current image ($N_\mathcal{T}$), were also examined. The results are shown in Fig. \ref{Fig: error_n1}. Each value in the figure was averaged over 100 repeated simulations, and then over six target images (5th to 10th).

From Fig. \ref{Fig: error_n1}a it was found that the classifier prediction procedure provided better result if previous classifiers had higher accuracies. It was also found that the prediction result was better by using more previous classifiers. While it is better to use a larger number of previous classifiers, in general, other factors, such as how good each previous classifier is and how long the period of the previous images cover, need to be considered for the selection of $N$.

After fine-tuning, classification accuracy of the current image was improved as shown in Fig. \ref{Fig: error_n1}b. The fine-tuned classifier of the current image provided comparable or better classification results than previous classifiers, especially when the previous classifiers were of low accuracies (60$\%$$\sim$70$\%$). This indicates that the proposed approach is able to compensate for errors in previous classifiers from the use of the training samples of the current image. It was also found that, previous classifiers of higher accuracies led to better classification accuracy of the current image. This phenomenon was more obvious when $N_\mathcal{T}$ = 5 than $N_\mathcal{T}$ = 50, suggesting that the accuracy of previous classifiers is more critical when the training samples of the current image is deficient. Therefore, when previous classifiers are of low accuracies, increasing the number of training samples of the current image should be an effective solution for achieving a better classifier training result.

\subsubsection{Stableness of Sequential Classifier Training with Previous Classifiers of Low Accuracies}

\quad

\begin{figure}[!htb]
\centering
\includegraphics[width=3.49in]{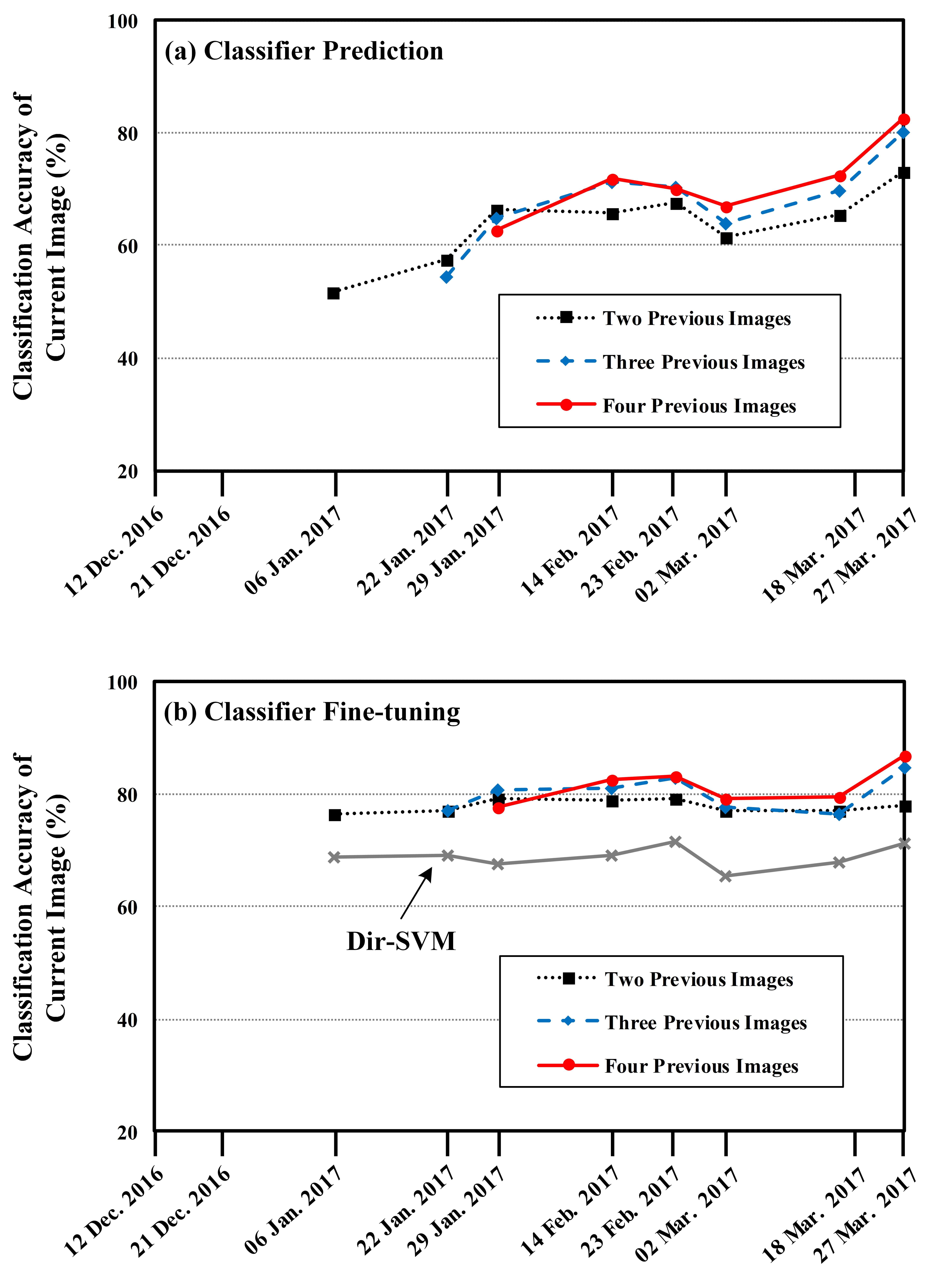}
\caption{Accuracies of sequential classifier training with previous images of low accuracies.}\label{Fig: error_n2}
\end{figure}

In this experiment, the insensitive of sequential classifier training to inaccurate initial set of previous classifiers was tested. We started with setting the first two, three, or four images as previous images, and their classifiers were selected from the 60$\%$$\sim$70$\%$ group. Then classifiers for the rest images (target images) were predicted and fine-tuned in a sequential order. For each target image, 50 labelled samples per class were selected. This process was repeated 100 times with the average being calculated. Fig. \ref{Fig: error_n2} shows the classification accuracies for the target images. It was found that, classifier prediction accuracies started with relatively low values, but improved gradually with time (Fig. \ref{Fig: error_n2}a). This indicates that unreliable previous classifiers will become less influential if they are more distant back in time with the current image. After classifier fine-tuning, the classification accuracies improved to around 80$\%$, higher than the original accuracies of 60$\%$$\sim$70$\%$, and remained relatively stable with time (Fig. \ref{Fig: error_n2}b). The prediction errors of the first a few target images (Fig. \ref{Fig: error_n2}a), affected by the poor previous classifiers, are largely compensated after fine-tuning as shown in Fig. \ref{Fig: error_n2}b. These results suggest that the proposed method is insensitive to the errors in the initial set of previous classifiers and can perform well after it is applied sequentially for a few times.

\section{Discussion and Conclusions}\label{sec: conclusion}

A SVM-based Sequential Classifier Training (SCT-SVM) approach is proposed for multitemporal remote sensing image classification. The approach leverages the classifiers of previous images to reduce the required number of training samples for the classifier training of a new image. Based on the temporal trend of previous classifiers, classifiers are firstly predicted for the new image. Then, a newly developed domain adaptation algorithm, Temporal-Adaptive Support Vector Machines (TA-SVM), is used to fine-tune the predicted classifiers into more accurate positions. Experimental results showed that the TA-SVM outperformed state-of-the-art algorithms in classifier fine-tuning. It was found that, when training data are insufficient, the overall classification accuracy of the incoming image was improved from 76.18$\%$ to 94.02$\%$ with the proposed SCT-SVM, compared with those obtained without the assistance from previous images. This demonstrates that the leverage of \emph{a priori} information from previous images can provide advantageous assistance for the classification of later images.

The proposed approach is suitable for monitoring the quantitative change of a given class, for example, increasing or decreasing of areas, and changes of spatial distributions. Natural changes are often gradual changes, which lead to smooth transitions, especially when the time interval is short. Human activities or disasters, for example, logging or forest fire, often bring abrupt changes. These discontinued cases are not considered in this study as they often need to be handled by defining new classes. Solutions to these cases can be found in \cite{demir2013updating,demir2013classification}. Basically, these solutions incorporate active learning heuristics to correct significant changes in class distributions and discover new classes. When abrupt changes occur, the proposed sequential classifier training needs to stop, and restart after a few previous images are available.

This study suggests that the assistance from historical image classification is able to ease the training data requirements for SVM-based classifications. This concept can be applied and extended to other classifiers, though specific algorithms need to be developed to accommodate the characteristics of those classifiers. For example, Gaussian classifiers need sufficient training data to make good estimation of class distributions. The parameters of class distributions can be predicted from those of historical images, and utilized to help the classifier training of an incoming image. Similar approaches for some other classifiers, such as random forests and neural networks, are less straight forward and need more analyses.

\section*{Acknowledgment}
The authors would like to thank Mr. Chris Quirk and Mr. Andrew Law at the Sunrice Milling Company for providing the rice distribution map, and Mr. Graham Parton and Mr. Bernard Star at the Coleambally Irrigation Co-operative Ltd for providing the farmland boundary data.

\bibliography{mybibfile}
\bibliographystyle{IEEEtran}

\end{document}